\documentclass{article}

\usepackage{microtype}
\usepackage{graphicx}
\usepackage{subcaption}
\usepackage{booktabs} 

\usepackage{hyperref}

\usepackage[accepted]{icml2026}

\usepackage{amsmath}
\usepackage{amssymb}
\usepackage{mathtools}
\usepackage{amsthm}
\usepackage[normalem]{ulem}
\usepackage[table]{xcolor}
\usepackage{multirow}
\usepackage{caption}

\usepackage{float}

\newlength\savewidth\newcommand\shline{\noalign{\global\savewidth\arrayrulewidth
  \global\arrayrulewidth 1pt}\hline\noalign{\global\arrayrulewidth\savewidth}}
\newcommand{\tablestyle}[2]{\setlength{\tabcolsep}{#1}\renewcommand{\arraystretch}{#2}\centering\footnotesize}

\newcolumntype{*}{>{\global\let\currentrowstyle\relax}}
\newcolumntype{^}{>{\currentrowstyle}}

\usepackage[capitalize,noabbrev]{cleveref}

\theoremstyle{plain}

\theoremstyle{definition}

\theoremstyle{remark}

\icmltitlerunning{CORE}

\begin{document}

\twocolumn[
  \icmltitle{CORE: Conflict-Oriented Reasoning for General Multimodal \\Manipulation Detection}



  \icmlsetsymbol{cor}{$\dagger$}

  \begin{icmlauthorlist}
    \icmlauthor{Jinjie Shen}{hfut,whu,lion}
    \icmlauthor{Yaxiong Wang}{cor,hfut,lion}
    \icmlauthor{Yujiao Wu}{csiro}
    \icmlauthor{Lechao Cheng}{hfut,lion}\\
    \icmlauthor{Tianrui Hui}{hfut,lion}
    \icmlauthor{Nan Pu}{hfut,lion}
    \icmlauthor{Zhihui Li}{ustc}
    \icmlauthor{Zhun Zhong}{cor,hfut,lion}
  \end{icmlauthorlist}

  \icmlaffiliation{hfut}{School of Computer Science and Information Engineering, Hefei University of Technology, Hefei, China}
  \icmlaffiliation{whu}{Wuhan University, Wuhan, China}
  \icmlaffiliation{lion}{Lab for Intelligence and visiON (LION)}
  \icmlaffiliation{csiro}{CSIRO, Australia}
  \icmlaffiliation{ustc}{University of Science and Technology of China, Hefei, China}

  \icmlcorrespondingauthor{Jinjie Shen}{shenjinjie22@gmail.com}
  \icmlcorrespondingauthor{Yaxiong Wang}{wangyx@hfut.edu.cn}
  \icmlcorrespondingauthor{Zhun Zhong}{zhunzhong007@gmail.com}

  \icmlkeywords{Forgery Detection, Multimodal, Conflict Reasoning, MLLM, Manipulation Detection}

  \vskip 0.3in
]



\makeatletter
\renewcommand{\printAffiliationsAndNotice}[1]{\global\icml@noticeprintedtrue%
  \stepcounter{@affiliationcounter}%
  {\let\thefootnote\relax\footnotetext{\hspace*{-\footnotesep}\ificmlshowauthors #1\fi%
      \forloop{@affilnum}{1}{\value{@affilnum} < \value{@affiliationcounter}}{
        \textsuperscript{\arabic{@affilnum}}\ifcsname @affilname\the@affilnum\endcsname%
          \csname @affilname\the@affilnum\endcsname%
        \else
          {\bf AUTHORERR: Missing \textbackslash{}icmlaffiliation.}
        \fi
      }.%
      \ificmlshowauthors
        { }First author: Jinjie Shen \textless{}shenjinjie22@gmail.com\textgreater{}. Corresponding to Yaxiong Wang~\mbox{\textless{}wangyx@hfut.edu.cn\textgreater{}}, Zhun Zhong~\mbox{\textless{}zhunzhong007@gmail.com\textgreater{}}.
      \fi
      \ \\
      \Notice@String
    }
  }
}
\makeatother
\printAffiliationsAndNotice{\textsuperscript{$\dagger$}Corresponding author.}

\begin{abstract}
The rapid rise of generative AI has made multimodal fake news increasingly realistic and pervasive, posing severe threats to public trust and social stability. Existing detection methods rely heavily on manipulation-specific models and large-scale labeled data, resulting in poor generalization to emerging manipulation types. We observed that the essence of manipulated misinformation lies in its intrinsic conflicts, \textbf{i.e.,} semantic or physical inconsistencies either across modalities or with common world knowledge. Inspired by this observation, we propose  \textbf{C}onflict-\textbf{O}riented \textbf{RE}asoning (\textbf{CORE}) framework, an effective paradigm that learns to endows multimodal large language models (MLLMs) with explicit conflict-capturing capability. To this end, CORE first constructs the Conflict Attribution Corpus (CAC) with fine-grained annotations of conflict factors and sources, providing essential data support for subsequent conflict perception training. By performing conflict-oriented representation enhancement and reasoning based on CAC, CORE achieves robust and generalizable conflict detection, effectively and rapidly adapting to unseen manipulation types with a few samples or in even zero-shot settings. Extensive experiments demonstrate that CORE surpasses state-of-the-art models. The dataset and code are publicly available at \url{https://github.com/shen8424/CORE}.
\end{abstract}

\section{Introduction}
\label{sec:intro}

The rapid advancement of generative artificial intelligence is profoundly impacting multiple domains~\cite{intro1,intro2,intro3,intro04}, deeply blurring the boundary between reality and fiction. In social network, malicious actors can now create highly convincing multimodal fake news, combining manipulated images with deceptive text, at an unprecedented scale and speed~\cite{intro05,intro06,intro07,intro08,intro4}. These forgeries, ranging from subtle edits of facial attributes to entirely fabricated scenes, pose a serious threat to public trust, social stability~\cite{intro4,intro5,intro6}. As manual verification becomes increasingly difficult, the development of robust automated detection systems is more critical than ever.
\begin{figure}
    \centering
    \includegraphics[width=\linewidth]{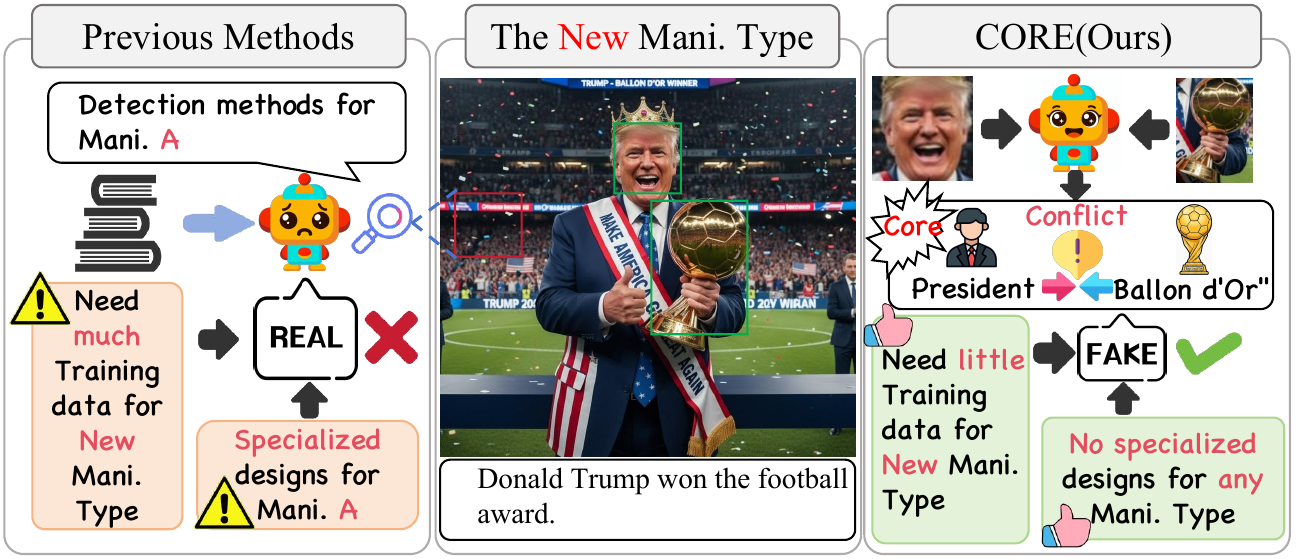}
    \caption{While previous methods require extensive data and specialized designs for specific manipulations, they struggle with new types. Our CORE addresses the core ``conflict'' in fake news, enabling generalized detection and excellent performance with minimal data. ``Mani.'' means ``Manipulation''}
    \label{fig:teaser}
\vspace{-0.6cm}
\end{figure}

In response to these challenges, various manipulation detection methods have been developed~\cite{dgm4,samm,mdsm,fka-owl,hammer++,clc}, successfully alleviating the rampant spread of multimodal fake news. However, the success of these methods is predicated on designing models and training paradigms tailored to specific manipulation types and relying on large-scale, type-specific training data. In practice, there is a continuous ``arms race'' between forgery techniques~\cite{simswap,hfgi,styleclip,infoswap} and detection methods. The evolutionary pace of new manipulation methods far outpaces both the cycle of data collection, cleaning, and annotation, and the targeted model design required for each novel type. As a result, current methods significantly degrades when encountering new manipulation patterns~\cite{yaxiongAIGCDet1,yaxiongAIGCDet2,yaxiongAIGCDet3}. 
Therefore, the field urgently requires a new paradigm that can move beyond dependencies on data and specific model designs, enabling models to achieve effective adaption with only a few samples to novel manipulation types~\cite{intro11,intro12,intro13}.

We observed that the essence of manipulated information lies in its intrinsic ``conflict''. This conflict can manifest as: a semantic contradiction between the \textbf{content} and \textbf{world knowledge}, such as the common-sense conflict between Trump's presidential status and football award in the news ``Donald Trump wins the football award''; 
or a conflict at the physical level, such as lighting and shadows, between the manipulated content and the original image/text. As shown in Figure~\ref{fig:teaser}, existing methods implicitly capture such conflict with massive training data and specialized model designs, but this over-reliance leads to overfitting towards specific manipulation patterns, resulting in poor generalization for new types. In contrast, humans detect deception by activating their knowledge and performing conflict reasoning, enabling robust judgment across diverse manipulation forms. Motivated by this consideration, we argue that \textit{if a detection model is endowed with explicit conflict-capturing capability, it can emulate human-like robustness when facing novel manipulation scenarios, thereby alleviating the long-standing data dependence and design rigidity of current approaches.}

\begin{figure*}[t]\centering
\centering\subfloat[Multimodal feature distribution of Qwen2.5VL-3B.
\label{fig:tsnebeforetrain}
]{\begin{minipage}[b]{0.95\linewidth}
\includegraphics[width=0.95\linewidth]{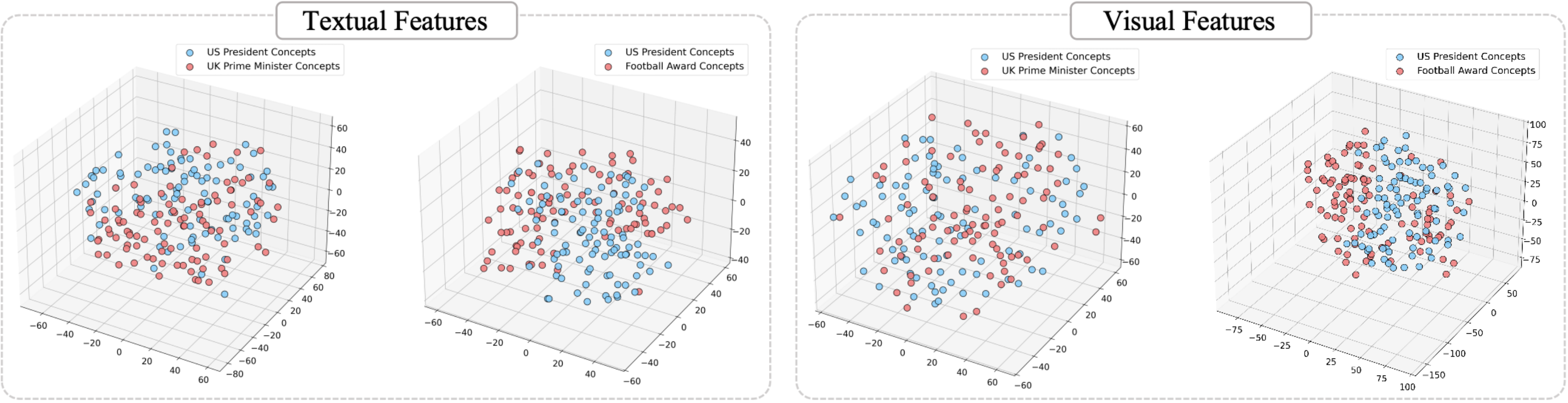}
\end{minipage}
}
\\
\subfloat[Multimodal feature distribution of $\text{CORE}_\text{Qwen}$(2.5VL-3B).
\label{fig:tsneaftertrain}
]{\begin{minipage}[b]{0.95\linewidth}
\includegraphics[width=0.95\linewidth]{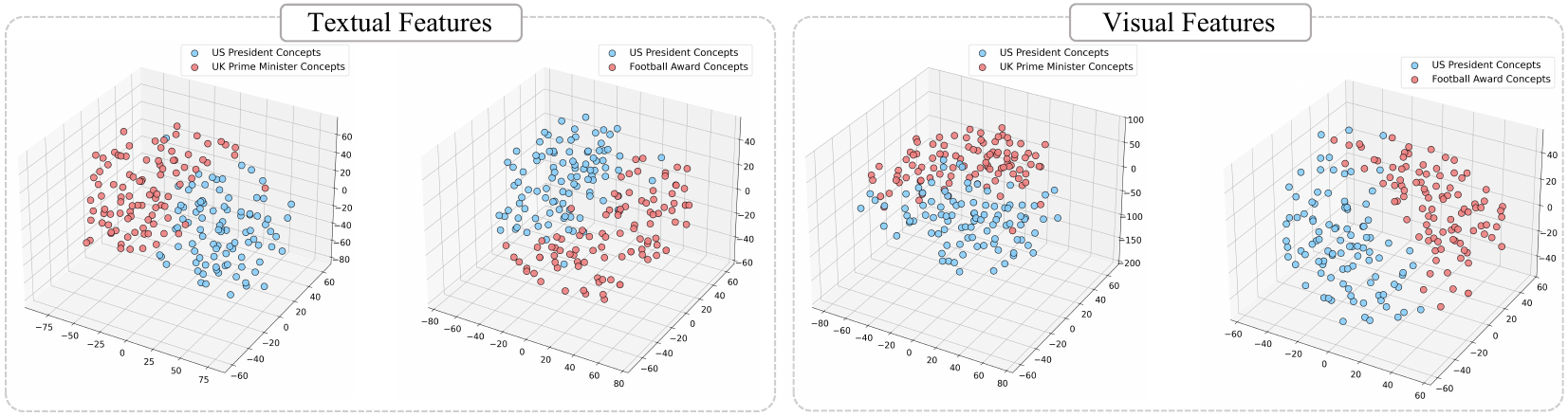}
\vspace{-0.1cm}
\end{minipage}
}
\vspace{-.5em}
\caption{Multimodal feature visualization of two group of conceptions from Qwen2.5VL-3B (a) and Qwen2.5VL-3B equipped with our CORE (b), where the textual and visual features are respectively shown from left to right.
}
\label{ablation}
\end{figure*}

Following the human reasoning process in detecting manipulations in multimodal misinformation, the ability to capture multimodal conflicts largely depends on a model's understanding of real-world knowledge. Multimodal Large Language Models (MLLMs)~\cite{qwen2.5vl,gemma3,llama3.2,seedvl}, trained on vast multimodal corpora, inherently encode rich world knowledge and thus exhibit strong potential for identifying conflicts in multimodal manipulations. However, they still fall short in conflict capturing due to the lack of conceptual understanding. MLLMs often conflate entirely unrelated concepts in the feature space, such as \textit{``U.S. President''} and \textit{``Football Award''} (Sec.~\ref{sec:3} Figure~\ref{fig:tsnebeforetrain}). Owing to this weakness, 
existing MLLMs, despite their rich world knowledge, still struggle to achieve robust and generalizable misinformation detection (Table~\ref{tab:0.1k-2.5k}).

To overcome the aforementioned limitations and establish a foundation model for general multimodal manipulation detection, we propose the Conflict-Oriented REasoning (CORE) framework. This framework equips MLLMs with explicit conceptual understanding, thereby enabling conflict detection capabilities. Training this capability requires explicit, fine-grained conflict supervision, which existing datasets lack. To provide this necessary data support, we first construct the Conflict Attribution Corpus (CAC). Each sample in CAC is annotated with both a conflict factor revealing the specific contradictory content within the misinformation and a conflict source, indicating whether the contradiction arises from the text, image, or underlying world knowledge. With these fine-grained annotations, we perform a Conflict-Perception Training (CPT) to perceive multimodal conflicts by enhancing the boundaries between conflicting concepts in the feature space, acquiring human-like conflict comprehension and detection ability.

With the acquired conflict-capturing capability from CPT, our CORE framework enables rapid adaptation to emerging manipulation patterns. Superior detection performance can be achieved with only a few-sample fine-tuning of new manipulation types, and even under zero-shot settings.
In summary, our main contributions are as follows:

\textbf{(1)}  We introduce an effective learning paradigm for general multimodal manipulation detection, which can rapidly adapt to novel manipulations with limited target samples. 

\textbf{(2)} Moving beyond the conventional paradigm of designing models for specific manipulations, we propose CORE, a general framework for multimodal misinformation detection that endows MLLMs with human-like conflict reasoning and enables fast adaptation to unseen misinformation patterns. 

\textbf{(3)} We construct the Conflict Attribution Corpus (CAC), a carefully curated dataset containing 14k samples with fine-grained annotations of conflict factors and sources, providing a solid benchmark for studying conflict reasoning in multimodal manipulation.

\begin{table*}[t]
\caption{World Knowledge Evaluation of Non-MLLMs and MLLMs.}
\label{tab:world_knowledge}
\centering

\subfloat[Average world knowledge evaluation, comparing Non-MLLMs (CLIP, ALBEF) against MLLMs (Qwen2.5VL-3B, Gemma3-4B).
\label{tab:world_knowledge_benchmark}
]{
    \begin{minipage}[t]{0.48\linewidth}
    \centering
    \begin{tabular}{l c}
        \toprule
        \textbf{Models} & \textbf{World Knowledge (ACC \%)} \\
        \midrule
        Non-MLLMs & 41 \\
        MLLMs & \textbf{96} \\
        \bottomrule
    \end{tabular}
    \end{minipage}
}
\hfill
\subfloat[Linear classification accuracy. Results are averaged across textual and visual modalities.
\label{tab:linear_separability}
]{
    \begin{minipage}[t]{0.48\linewidth}
    \centering
    \begin{tabular}{l c} 
        \toprule
        \textbf{Classification Task} & \textbf{ACC (\%)} \\
        \midrule
         Pres. vs. Football Award & 61 \\
         Pres. vs. UK Prime Minister & 53 \\
        \bottomrule
    \end{tabular}
    \end{minipage}
}
\end{table*}

\noindent\textbf{Conflict of Interest Disclosure.} The authors declare no conflicts of interest.

\section{Related Works}

As deepfake technologies continue to evolve, research in the field of multimodal disinformation detection has also made significant progress. For instance, models such as HAMMER~\cite{dgm4} and ASAP~\cite{asap} have designed specialized contrastive learning and fine-grained detection modules to address the specific problem of image-text inconsistency; meanwhile, RamDG~\cite{samm} focuses on celebrity-related fake news, employing external knowledge bases for targeted detection. In recent years, the rise of MLLMs has pushed research to new heights. SNIFFER~\cite{sniffer} designed a specialized two-stage fine-tuning process to enhance the ability to judge image-text consistency, while FKA-Owl~\cite{fka-owl} attempts to tackle specific types of common sense fallacies by integrating world knowledge. To handle more complex forgeries, MMD-Agent~\cite{mmfakebench} constructs a specific multi-step reasoning framework, and AMD~\cite{mdsm} relies on detailed prior information, such as manipulation region coordinates and manipulation types, for detection.

Despite the considerable progress in multimodal news detection methods, they suffer from two limitations. First, they heavily rely on large-scale datasets constructed for specific manipulation types; second, their model designs or training strategies are often specialized for certain forgery traces. These specialized designs make it difficult to guarantee the models generalization ability when faced with out-of-distribution, and especially unseen, manipulation types. Therefore, our work moves away from designing for specific forgery traces and instead focuses on the core flaw of forged information—conflict. Mastering this fundamental capability allows the model to break its dependence on large-scale, specific data, thereby demonstrating robust generalization and detection capabilities for unseen manipulation types in a few-sample and even zero-shot scenarios.

\begin{figure*}
    \centering
    \includegraphics[width=1\textwidth]{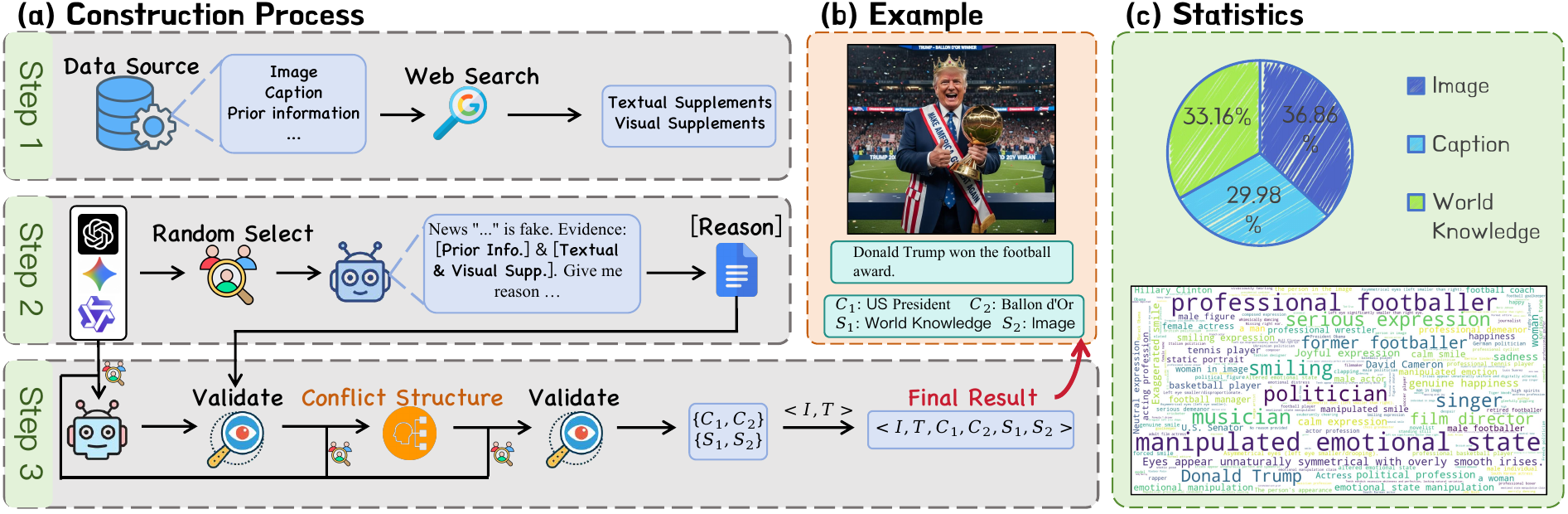}
    \caption{(a) The construction process of CAC. (b) An example from CAC. (c) Statistics of CAC, including the distribution of conflict sources and word clouds of the Conflict Factor.}
    \label{fig:Dataset}
\end{figure*}

\section{The Challenge of Conflict Perception}
\label{sec:3}
Human often do not require extensive training on similar samples to identify novel forged information. This is largely attributable to their ability to acutely identify conflicts within news based on their world knowledge and understanding. This capability is built upon two core foundations: 1) a comprehensive repository of world knowledge, and 2) a clear and well understanding of that knowledge to support conflict-capturing. This section investigates through a series of experiments whether current mainstream models possess these two key capabilities.

We first investigate a fundamental question: \textit{Do existing models possess the world knowledge required to identify fake news?} To this end, we constructed a benchmark of 200 multiple-choice questions, covering the diverse world knowledge needed to detect fake news (See Appendix~\ref{sec:wk_benchmark} for details). Our evaluation includes two representative classes of models: non-MLLM models, such as CLIP~\cite{clip} and ALBEF~\cite{albef}, and MLLMs, including Qwen2.5VL-3B and Gemma3-4B. For the non-MLLM models, we assess their choices by calculating the cosine similarity between the embeddings of the question and the options after they pass through the encoder; the option with the highest similarity is considered the model's prediction. For MLLMs, we directly use prompting to have them output the correct option. The experimental results, as shown in Table~\ref{tab:world_knowledge_benchmark}, indicate that MLLMs possess relatively complete knowledge, whereas non-MLLMs do not.

To investigate whether MLLMs possess clear conceptual boundaries like humans, we systematically analyze their feature representation space. We select concept pairs with varying semantic differences (\textit{e.g.}, U.S. President vs. football player; U.S. President vs. UK Prime Minister), collect 100 relevant entities for each concept, and extract their multimodal features (See the Appendix~\ref{appendixc-feature_extraction} for details). We then use t-SNE~\cite{tsne} for visualization. As shown in Figure~\ref{fig:tsnebeforetrain}, the results demonstrate that the MLLM's representation space fails to form clear boundaries: the distributions for even semantically disparate concepts are diffuse and overlapping; We further train a classifier based on the features to quantify their separability, the low classification accuracy in Table~\ref{tab:linear_separability} also quantitatively confirms this.

The experiments show that non-MLLM models suffer from incomplete knowledge, while MLLMs, though addressing the knowledge repository issue, still lack conceptual clarity. The key to resolving this dilemma is to enable models to possess knowledge and understand it in a clear, structured way, thereby learning detection based on the core principle of conflict.

\section{Methodology}
\begin{figure*}
    \centering
    \includegraphics[width=1\textwidth]{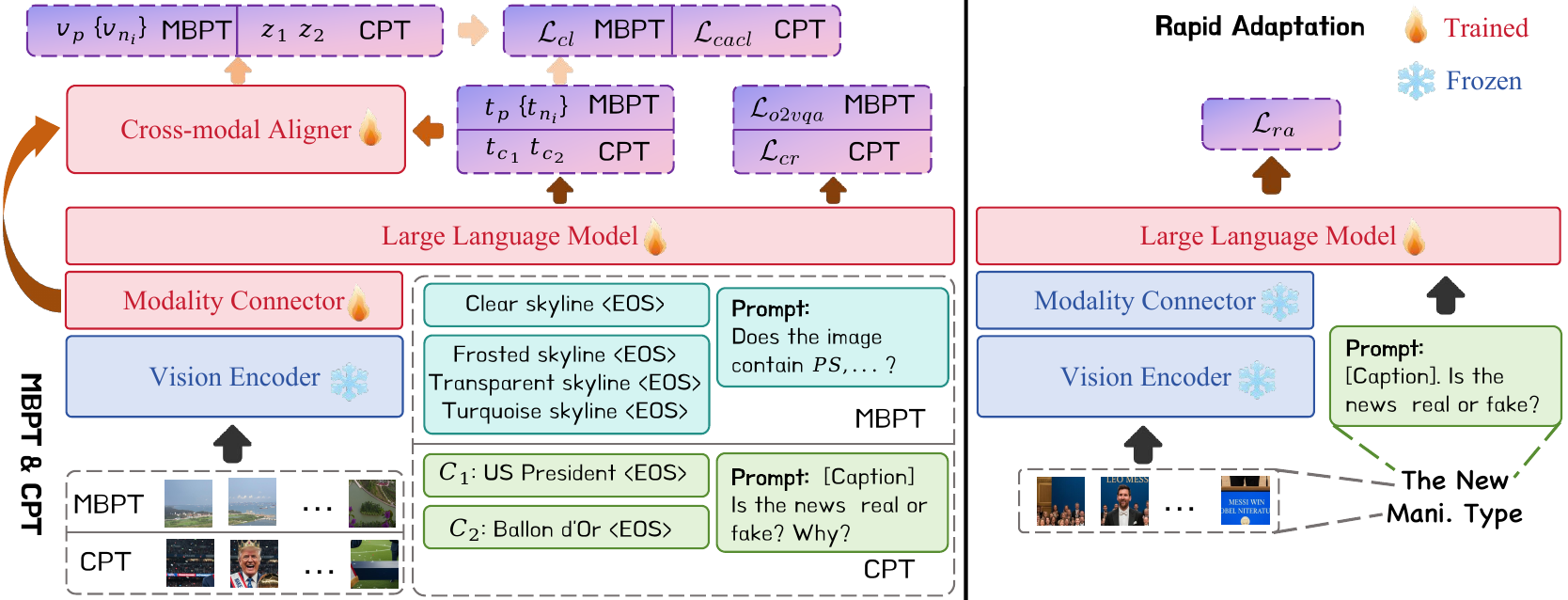}
    \caption{The architecture of our proposed CORE. It first employs MBPT to train cross-modal alignment, subsequently utilizes CPT to train conflict perception, and finally achieves effective detection of novel manipulations via Rapid Adaptation.}
    \label{fig:csara}
\vspace{-0.3cm}
\end{figure*}

To overcome the above limitations, we propose the Conflict-Oriented Reasoning (CORE) framework, as illustrated in Figure~\ref{fig:csara}. Built upon MLLMs, CORE leverages their extensive knowledge base and reshapes the boundaries between conflicting concepts to enhance the model's conceptual understanding, thereby improving conflict perception capability and enabling MLLMs with human-like conflict capturing ability. 
To enable this, we first construct the Conflict Attribution Corpus (CAC) with fine-grained annotations of conflict sources and factors. Next, Modality Bridging Pre-Training (MBPT) is conducted to train a Cross-Modal Aligner. This aligner bridges the modality gap, enabling the full utilization of CAC annotations. Finally, the Conflict Perception Training (CPT) stage explicitly reshapes the model's conceptual understanding of conflicting elements, thereby refining its ability to perceive and reason over multimodal conflicts.

\subsection{Conflict Attribution Corpus}
The CAC provides explicit supervision signals to facilitate conflict perception learning.
As illustrated in Figure~\ref{fig:Dataset}(a), the construction of CAC involves the following steps:

\noindent\textbf{-Source Sample Selection.} Given that the SAMM~\cite{samm} provides rich manipulation annotations including manipulated objects and regions, which offer valuable cues for subsequent conflict attribution generation. We therefore select 100k image-text pairs from it as our base data.

\noindent\textbf{-Background Knowledge Collection.} Next, we collect external supplementary materials that are semantically related to each image–text pair as background knowledge via the Google Search API~\cite{googleapi}, providing reliable and comprehensive support for conflict reasoning.

\noindent\textbf{-Conflict Rationale Generation}. Subsequently, the integrated information including the multimodal inputs, manipulation prior and the background information are fed into a MLLM randomly selected from an expert pool of \{GPT-4o, Gemini2.5-Pro, Qwen3-VL-Plus\}~\cite{gpt4o,gemini,qwen3vlplus}, instructing it to generate a detailed reason why the news is false. The conflict rationale is then cross-validated for plausibility by the other two MLLMs.

\noindent\textbf{-Conflict Structuring}. After validation, the sample is once again sent to a randomly selected MLLM to distill the reason into a structured format of {Conflict Factor 1, Conflict Factor 2} and their respective {Conflict Source 1, Conflict Source 2}, where the conflict factor specifies the content of the contradiction, while the conflict source pinpoints its origin, as shown in Figure~\ref{fig:Dataset}(b). This result also undergos a final review by the remaining two MLLMs.

\noindent\textbf{Statistics.}
As shown in Figure~\ref{fig:Dataset}(c), CAC contains 14k instances. Its final data structure is \textless Image $I$, Text $T$, \{Conflict Factor $C_1$, Conflict Factor $C_2$\}, \{Conflict Source $S_1$, Conflict Source $S_2$\}\textgreater. Regarding the distribution of conflict sources, 29.98\% of conflicts originate from the news caption, 36.86\% originate from the news image, and 33.16\% from world knowledge. Please refer to Appendix~\ref{sec:appendix_prompts} for the prompts and validation protocols.

\definecolor{LightGreen}{rgb}{0.88, 1, 0.88}
\definecolor{Red}{rgb}{1, 0, 0}

\begin{table*}
\centering
\setlength{\tabcolsep}{3pt}
\caption{
Performance comparison (ACC) on multiple datasets using 100-750 (100-350) samples.
}
\scalebox{0.95}{\begin{tabular}{l cccc cccc cccc cccc}
\toprule
\multirow{2}{*}{\textbf{Method}} & \multicolumn{4}{c}{\textbf{DGM$^4$}} & \multicolumn{4}{c}{\textbf{MDSM}} & \multicolumn{4}{c}{\textbf{MMFakeBench}} & \multicolumn{4}{c}{\textbf{NewsCLIPpings}} \\
\cmidrule(lr){2-5} \cmidrule(lr){6-9} \cmidrule(lr){10-13} \cmidrule(lr){14-17}

& 100 & 200 & 500 & 750 & 100 & 200 & 500 & 750 & 100 & 200 & 300 & 350 & 100 & 200 & 500 & 750 \\
\midrule
Qwen3VL-235B & \underline{51.6} & \underline{51.6} & 51.6 & 51.6 & 56.2 & 56.2 & 56.2 & 56.2 & 57.4 & 57.4 & 57.4 & 57.4 & 61.3 & 61.3 & 61.3 & 61.3 \\
Gemma3-27B & 49.3 & 49.3 & 49.3 & 49.3 & 52.9 & 52.9 & 52.9 & 52.9 & 53.7 & 53.7 & 53.7 & 63.7 & 58.8 & 58.8 & 58.8 & 58.8 \\
LLaMA3.2-90B & 48.3 & 48.3 & 48.3 & 48.3 & 50.8 & 50.8 & 50.8 & 50.8 & 54.1 & 54.1 & 54.1 & 54.1 & 57.9 & 57.9 & 57.9 & 57.9 \\
SeedVL-1.5 & 50.8 & 50.8 & 50.8 & 50.8 & \underline{57.2} & \underline{57.2} & 57.2 & 57.2 & 60.4 & 60.4 & 60.4 & 60.4 & \underline{62.4} & \underline{62.4} & \underline{62.4} & \underline{62.4} \\
\midrule 

HAMMER & 45.3 & 49.8 & 53.8 & 56.9 & 48.1 & 53.9 & 59.8 & 62.2 & 59.7 & 64.7 & 67.1 & 68.2 & 54.4 & 56.2 & 57.4 & 57.3 \\
HAMMER++ & 45.8 & 49.8 & 53.7 & 57.1 & 48.0 & 54.0 & \underline{60.0} & 62.2 & 59.5 & \underline{64.9} & 67.0 & \underline{68.3} & 54.5 & 56.0 & 57.4 & 57.3 \\
RamDG & 46.4 & 50.3 & \underline{55.0} & 57.9 & 48.9 & 53.7 & 58.7 & 61.7 & 60.9 & 64.4 & \underline{68.1} & 68.2 & 55.3 & 56.9 & 57.1 & 57.3 \\
\midrule 
FKA-Owl & 47.4 & 51.0 & 51.2 & 52.4 & 41.3 & 43.8 & 45.3 & 51.1 & 49.9 & 53.2 & 58.4 & 59.1 & 45.4 & 48.0 & 49.8 & 50.0 \\
AMD & 37.9 & 40.5 & 40.7 & 40.8 & 38.2 & 40.7 & 43.9 & 51.3 & 48.3 & 51.4 & 53.9 & 55.2 & 40.4 & 42.1 & 45.6 & 46.7 \\
Qwen2.5VL-3B & 48.0 & 49.7 & 53.4 & 56.3 & 50.3 & 53.2 & 58.1 & 60.3 & 60.2 & 62.4 & 65.3 & 66.5 & 49.8 & 51.9 & 57.2 & 60.4 \\
Gemma3-4B & 48.3 & 50.1 & 54.3 & \underline{60.8} & 49.8 & 52.0 & 57.7 & \underline{63.0} & \underline{61.1} & 64.3 & 66.4 & 66.7 & 48.9 & 51.0 & 57.0 & 61.4 \\
\midrule 

\rowcolor{LightGreen}
\textbf{$\text{CORE}_\text{Qwen}$} & 59.7 & \textbf{63.9} & \textbf{65.2} & 65.4 & \textbf{69.0} & \textbf{70.1} & 74.1 & 74.5 & 73.5 & 76.4 & \textbf{79.4} & \textbf{79.4} & \textbf{64.3} & \textbf{70.7} & \textbf{70.9} & \textbf{71.0} \\
\rowcolor{LightGreen}
\textbf{$\text{CORE}_\text{Gemma}$} & \textbf{61.3} & 61.9 & 65.2 & \textbf{68.4} & 68.4 & 70.0 & \textbf{79.3} & \textbf{82.0} & \textbf{75.2} & \textbf{76.7} & 78.1 & 78.1 & 63.0 & 68.4 & 69.6 & 70.9 \\
\midrule

\textbf{$\Delta$ (vs Best Baseline)} & 
{\color{Red} +9.7} & 
{\color{Red} +12.3} & 
{\color{Red} +10.2} & 
{\color{Red} +7.6} & 
{\color{Red} +11.8} & 
{\color{Red} +12.9} & 
{\color{Red} +19.3} & 
{\color{Red} +19.0} & 
{\color{Red} +14.1} & 
{\color{Red} +11.8} & 
{\color{Red} +11.3} & 
{\color{Red} +11.1} & 
{\color{Red} +1.9} & 
{\color{Red} +8.3} & 
{\color{Red} +8.5} & 
{\color{Red} +8.6} \\ 
\bottomrule
\end{tabular}}
\small
\label{tab:0.1k-2.5k}
\vspace{-0.1cm}
\end{table*}
\subsection{Modality Bridging Pre-Training}
Although the sources of conflict span both visual and textual modalities, their annotated form in CAC is uniformly text. Therefore, accurately mapping conflict descriptions that originate from vision but exist in text form back to the visual space becomes a challenge. To bridge this modality gap, we introduce a concise and efficient Cross-modal Aligner and forge its cross-modal alignment capability through a dedicated pre-training stage. 
After the model acquires reliable alignment capabilities, we then commence the second stage of conflict perception training on CAC.

This stage of training is conducted on 50k samples from the FineHARD~\cite{finehard} dataset, whose samples consist of an image~$I$, a positive sample $\mathit{PS}$ that exists in the image, and three hard negative samples~$\{\mathit{NS}_i\}_{i=1}^3$ that are semantically close to $\mathit{PS}$ but do not exist in the image (see Appendix for examples). For feature extraction, the image~$I$ is passed through the vision encoder~$\mathcal{E}_V$ and a modality connector~$\mathcal{P}$ to obtain the visual feature sequence~$V = \mathcal{P}(\mathcal{E}_V(I))$. Simultaneously, the positive and negative text samples, each appended with an \texttt{<EOS>} token, are fed into the LLM. We then extract the hidden state corresponding to the \texttt{<EOS>} token from the LLM's final layer as the global feature.
From this, we get the global text feature for the positive sample $\mathbf{t}_{p}$, and a set of global text features for the negative samples,~$\{\mathbf{t}_{n_i}\}$. Next, using~$\mathbf{t}_{p},\{\mathbf{t}_{n_i}\}$~as the Query and~$V$~as the Key and Value~\cite{transformer}, we compute text-guided visual features~$\mathbf{v}_{p}, \{\mathbf{v}_{n_i}\}$~via a Cross-modal Aligner, which is simply implemented as a cross-attention layer:
\begin{equation}
\mathbf{v}_{p} = \operatorname{Aligner}\left(\mathbf{t}_{p}, V, V\right), \{\mathbf{v}_{n_i}\} = \operatorname{Aligner}\left(\{\mathbf{t}_{n_i}\}, V, V\right).
\label{eq:stage1_cross_attn}
\end{equation}

To achieve a fine-grained alignment, we adopt the contrastive learning loss proposed by SigLIP~\cite{siglip}, which aims to ensure that the extracted visual feature~$\mathbf{v}_{p}$~is semantically highly correlated with the corresponding text feature~$\mathbf{t}_{p}$:
\begin{equation}
\mathcal{L}_{cl} = \sum_{\mathbf{(t,v)} \in Q} \frac{1}{1 + e^{y_{\mathbf{t}}(s_1 \cdot \langle \mathbf{t}, \mathbf{v} \rangle + b_1)}},
\label{eq:contrastive_loss}
\end{equation}
where~$Q=\{(\mathbf{t}_{p},\mathbf{v}_{p}),\{(\mathbf{t}_{n_i},\mathbf{v}_{n_i})\},\{(\mathbf{t}_{n_i},\mathbf{v}_{p})\}\}$,
~$s_1$~and~$b_1$~are learnable scalar parameters, and~$\langle \cdot, \cdot \rangle$~represents cosine similarity. When~$(\mathbf{t},\mathbf{v})=(\mathbf{t}_{p},\mathbf{v}_{p})$~,~$y_{\mathbf{t}}=1$; otherwise,~$y_{\mathbf{t}}=-1$.

To aid fine-grained multimodal understanding and preserve the model's inherent language capabilities, we further devise an object-occurrence-based visual question answering task. Specifically, we construct the following question-answering~\cite{vqa} instruction format:

\noindent\texttt{\textbf{Question:}} ``Does the image contain $\text{RandSF}(\{PS, NS_1, NS_2,NS_3\})$?"

\noindent\texttt{\textbf{Answer:}} ``The image contains $\mathit{PS}$ and doesn't contain $\{\mathit{NS}_1,\mathit{NS}_2,\mathit{NS}_3\}$."

\noindent where RandSF($\cdot$) is the random shuffling operation. We then calculate a language generation loss~$\mathcal{L}_{o2vqa}$. The total loss function for this stage is defined as follows:
\begin{equation}
\mathcal{L}_{mbpt} = \mathcal{L}_{cl} + \mathcal{L}_{o2vqa}.  
\label{eq:stage1_loss}
\end{equation}

\begin{figure*}
    \centering
    \includegraphics[width=1\textwidth]{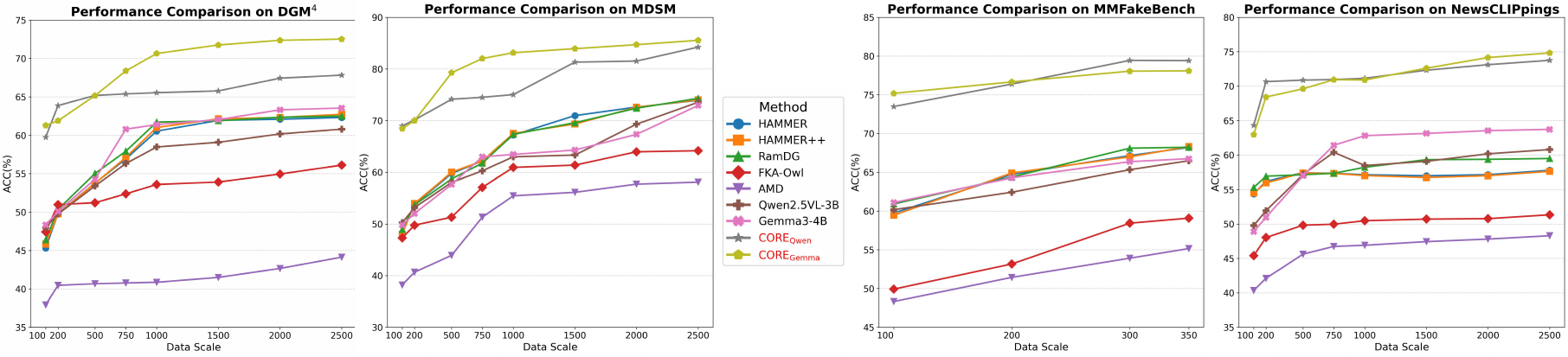}
    \vspace{-0.6cm}
    \caption{Performance comparison on multiple datasets using 100-2.5k (100-350 on MMFakeBench) samples.}
    \label{fig:0.1k-2.5k}
    \vspace{-0.5cm}
\end{figure*}

\subsection{Conflict Perception Training}
Given a sample~$<I, T, \{C_1, C_2\}, \{S_1, S_2\}> \in \text{CAC}$, the news image~$I$~is passed through~$\mathcal{E}_V$~and~$\mathcal{P}$~to obtain the visual feature sequence~$V$. The two conflict factors~$C_1, C_2$, after appending the~\texttt{<EOS>}~token, are fed into the LLM to obtain their corresponding global features ~$\mathbf{t}_{c_1}, \mathbf{t}_{c_2}$. We process the conflict factor features based on the modality of their sources~$\{S_1, S_2\}$. If the source~$S_i$~of a conflict factor~$C_i$~is the visual modality (i.e., $S_i$ is image), we invoke the Cross-
modal Aligner pre-trained in Modality Bridging Pre-Training (MBPT) stage to extract the corresponding visual feature. Otherwise, we keep the feature unchanged.
For ease of notation, we denote the two conflict representations used for comparison as~$\mathbf{z}_1$~and~$\mathbf{z}_2$, defined as follows:
\begin{equation}
\mathbf{z}_i = 
\begin{cases} 
\operatorname{Aligner}\left(\mathbf{t}_{c_i}, V, V\right) & \text{if } S_i \text{ is image,} \\
\mathbf{t}_{c_i} & \text{otherwise.}
\end{cases}
\end{equation}

First, we adopt a conflict-aware contrastive loss~$\mathcal{L}_{cacl}$~to help the model establish clear conceptual boundaries by pushing the two conflict factor representations~$\mathbf{z}_1$~and~$\mathbf{z}_2$~far apart in the semantic space, which is the core to identify conflicts in manipulated samples:
\begin{equation}
\mathcal{L}_{cacl} = \frac{1}{1 + e^{-(s_2 \cdot \langle \mathbf{z}_1, \mathbf{z}_2 \rangle + b_2)}}.
\label{eq:stage2_contrastive_loss}
\end{equation}
This loss function aims to maximize the distance between the two conflict factor representations.

Besides, we further design a conflict reasoning loss to enhance the conflict capture and preserve the model's inherent language capabilities:

\noindent\texttt{\textbf{Question:}} ``Does the news Real or Fake? If it's fake, further give the reason."

\noindent\texttt{\textbf{Answer:}} ``Real. / Fake. Because the $C_1$ from $S_1$ conflicts with $C_2$ from $S_2$."

We then calculate a language modeling loss~$\mathcal{L}_{cr}$ to produce conflict reasoning. The total loss function for CPT stage is defined as follows:
\begin{equation}
\mathcal{L}_{cpt} = \mathcal{L}_{cacl} + \mathcal{L}_{cr}.
\label{eq:stage2_loss}
\end{equation}


\subsection{Rapid Adaptation to Novel Manipulation}
To verify whether the model improved by CORE framework have a clear conceptual understanding, we re-examine the concept pairs of "US President vs. Football Award" and "US President vs. UK Prime Minister". As shown in  Figure~\ref{fig:tsneaftertrain}, the boundaries of conceptual embedding from the model become clear. The MLLMs with rich real-world knowledge and clear conceptual understanding hold good ability for conflict-capturing in manipulated multimodal misinformation.
Therefore, when facing with new manipulation types or news patterns, it only requires a small amount of data for fine-tuning to adapt quickly and achieve superior recognition performance. To ensure the model's generalization ability, we do not make specialized designs for specific data types. We only have the model make predictions by constructing a question-answering instruction, \textit{``Is the news real or fake?"}, and calculate the corresponding language generation loss~$\mathcal{L}_{ra}$ when finetuning. Similarly, during inference, the model requires only this simple instruction to make predictions.

\section{Experiments}

\begin{table}[t]
\centering
\setlength{\tabcolsep}{4pt} 
\caption{
Performance comparison (ACC) on large-scale data.
}
\begin{tabular}{l ccc}
\toprule
\textbf{Method} & \textbf{SAMM} & \textbf{MDSM} & \textbf{NewsCLIPpings} \\
\midrule

HAMMER & 92.26 & 86.09 & 64.22 \\
HAMMER++ & 92.43 & 86.31 & 63.95 \\
RamDG & 94.66 & 87.42 & 66.79 \\
\midrule 

FKA-Owl & 92.60 & 87.15 & 68.24 \\
AMD & 87.53 & 80.49 & 60.45 \\
Qwen2.5VL-3B & 92.49 & 87.23 & 71.43 \\
Gemma3-4B & 93.11 & 87.07 & 73.65 \\
\midrule 

\rowcolor{LightGreen}
\textbf{$\text{CORE}_\text{Qwen}$} & 96.74 & \textbf{88.49} & 81.87 \\
\rowcolor{LightGreen}
\textbf{$\text{CORE}_\text{Gemma}$} & \textbf{97.14} & 87.61 & \textbf{83.45} \\
\bottomrule
\end{tabular}
\small
\label{tab:large-scale}
\vspace{-0.4cm}
\end{table}

\noindent\textbf{Implementation Details.}
To validate the generalization ability of the proposed CORE framework, we select two advanced open-source MLLMs as our backbones: Qwen2.5VL-3B~\cite{qwen2.5vl} and Gemma3-4B~\cite{gemma3}, and apply the CORE framework to train them. All our training processes utilize the LoRA~\cite{lora} technique. Please refer to the Appendix~\ref{appendix-implementation_details} for more details.

\noindent\textbf{Datasets.}
To comprehensively evaluate the model's performance,  
we select four public multimodal datasets with diverse manipulation patterns: DGM$^4$~\cite{dgm4}, MDSM~\cite{mdsm}, MMFakeBench~\cite{mmfakebench}, and NewsCLIPpings~\cite{newsclippings}. To simulate the real-world scenario where novel manipulated data is scarce, we randomly sample a small number of samples from the aforementioned datasets for training. It is worth noting that, since MMFakeBench~\cite{mmfakebench} does not include a training set, we use its validation set as training data. Please refer to the Appendix~\ref{sec:appendix_data_overlap} for the data overlap analysis between FineHARD, CAC, and the benchmarks.

\noindent\textbf{Baselines.}
For a comprehensive comparison, we select various advanced Multimodal Manipulation Detection methods as baselines and compared their performance against $\text{CORE}_\text{Qwen}$ and $\text{CORE}_\text{gemma}$ on multiple datasets. The baseline models include non-MLLMs: HAMMER~\cite{dgm4}, HAMMER++~\cite{hammer++}, RamDG~\cite{samm}; as well as MLLMs specialized for detection tasks: FKA-Owl~\cite{fka-owl}, AMD~\cite{mdsm}, Qwen2.5VL-3B~\cite{qwen2.5vl}, and Gemma3-4B~\cite{gemma3}. In addition, we also introduce general-purpose MLLMs with larger parameter scales: Qwen3VL-235B~\cite{qwen3vlplus}, Gemma3-27B~\cite{gemma3}, LLaMA-3.2-Vision-90B~\cite{llama3.2}, and SeedVL-1.5~\cite{seedvl} for zero-shot.


\begin{table*}[h]
\centering\subfloat[Zero-shot performance comparison.
\label{fig:crotype}
]{\begin{minipage}[b]{0.52\linewidth}
\includegraphics[width=0.9\linewidth]{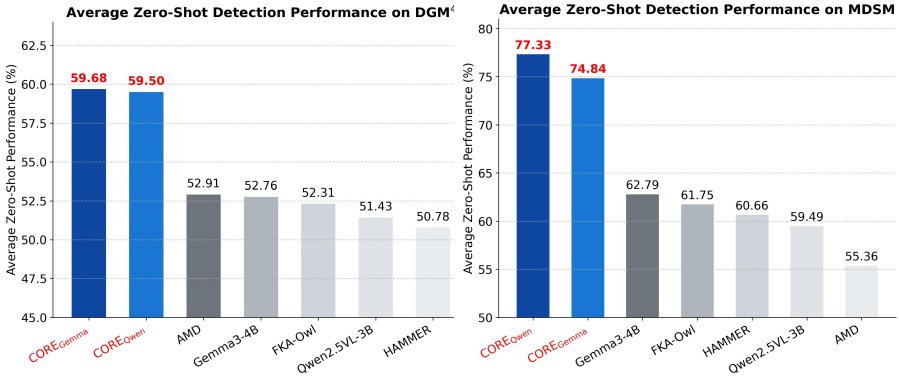}
\end{minipage}
}
\hspace{-0.5cm}
\subfloat[Data scale discussion in MBPT stage with $\text{CORE}_\text{Qwen}$.
\label{tab:mbpt_data}
]{
    \centering
    \begin{minipage}[l]
        {0.45\linewidth}{
            \begin{center}
            \vspace{-0.1cm}
            \tablestyle{4pt}{1.1}\footnotesize
\scalebox{0.93}{\begin{tabular}{l cccc}

\textbf{Volume} & \textbf{DGM$^4$} & \textbf{MDSM} & \textbf{MMFakeBench} & \textbf{NewsCLIPpings} \\
\shline

MBPT-0k & 60.42 & 69.95 & 75.44 & 65.96 \\
MBPT-25k & 63.13 & 71.02 & 77.42 & 68.19 \\

\rowcolor{LightGreen}
MBPT-\textbf{50k} & \textbf{65.18} & \textbf{74.13} & \textbf{79.44} & \textbf{70.88} \\

MBPT-75k & 65.12 & 74.03 & 79.48 & 70.92 \\
\end{tabular}}
            \end{center}
        }
        \vspace{0.4cm}
    \end{minipage}
}
\\
\vspace{0.2cm}
\hspace{-2em}
\subfloat[Impact of different loss in MBPT with $\text{CORE}_\text{Qwen}$.
\label{tab:mbpt_loss} 
]{
    \begin{minipage}
    {0.45\linewidth}{
        \begin{center}
        \vspace{-0.1cm}
        \tablestyle{4pt}{1.1}\footnotesize
\scalebox{0.93}{\begin{tabular}{l cccc}
\textbf{Loss Type} & \textbf{DGM$^4$} & \textbf{MDSM} & \textbf{MMFakeBench} & \textbf{NewsCLIPpings} \\
\shline

MSE & 62.52 & 71.03 & 76.42 & 68.97 \\

\rowcolor{LightGreen}
\textbf{Contrastive} & \textbf{65.18} & \textbf{74.13} & \textbf{79.44} & \textbf{70.88} \\
\end{tabular}}
        \end{center}
    }
    \vspace{0.01cm}
    \end{minipage}
}
\hspace{1.1cm}
\centering
\subfloat[Data scale discussion in CPT stage with $\text{CORE}_\text{Qwen}$.
\label{tab:cpt_data}
]{
    \centering
    \begin{minipage}
    {0.45\linewidth}{
    \vspace{-0.1cm}
        \begin{center}
        \tablestyle{4pt}{1.1}\footnotesize
\scalebox{0.93}{\begin{tabular}{l cccc}
\textbf{Volume} & \textbf{DGM$^4$} & \textbf{MDSM} & \textbf{MMFakeBench} & \textbf{NewsCLIPpings} \\
\shline

CPT-0K & 60.51 & 70.29 & 74.62 & 65.91 \\
CPT-7k & 64.74 & 73.20 & 78.96 & 69.57 \\

\rowcolor{LightGreen}
CPT-\textbf{14k} & \textbf{65.18} & \textbf{74.13} & \textbf{79.44} & \textbf{70.88} \\

CPT-38k & 64.93 & 73.63 & 78.92 & 70.42 \\
\end{tabular}}
        \end{center}
    }
    \end{minipage}
}
\vspace{0.2cm}
\caption{\textbf{Ablation and discussion experiments} of (a) average zero-shot performance comparison on $\text{DGM}^4$ and MDSM, (b) impact of data scale in MBPT stage, (c) impact of different loss types in MBPT stage, and (d) the impact of data scale in CPT stage.
}
\label{tab:test}
\vspace{-0.2cm}
\end{table*}

\subsection{Performance Comparison}

\noindent\textbf{Rapid Adaptation with a Few Samples.}
With the conflict-capturing ability of CORE framework, the model can achieve a rapid adaption to novel manipulations with limited data. To verify this, we construct training subsets by randomly sampling 100, 200, 500, 750, 1k, 1.5k, 2k, and 2.5k samples from each dataset, respectively. Table~\ref{tab:0.1k-2.5k} shows the performance of all methods with training set sizes of 100, 200, 500 and 750. Figure~\ref{fig:0.1k-2.5k} shows the performance trend as the training set size increases (100-2.5k). 

As shown in Table~\ref{tab:0.1k-2.5k} and Figure~\ref{fig:0.1k-2.5k}, both $\text{CORE}_\text{Qwen}$ and $\text{CORE}_\text{Gemma}$ achieve SOTA performance across multiple datasets when training data is limited, surpassing the second-best methods by an average of \textbf{9.94\%} and \textbf{10.50\%}, respectively. This demonstrates that the CORE framework can rapidly adapt to novel image manipulation techniques. Furthermore, the strong performance of CORE across two different MLLMs validates the method's generalization ability.


\noindent\textbf{Zero-shot Cross-Manipulation Detection.} 
To evaluate model's generalization, we utilize the MDSM and DGM$^4$ datasets, as both feature multiple manipulation categories, each with a sufficient number of training samples. In our experimental design, for each specific (target) manipulation type within a dataset, we train a model using 1k randomly sampled instances composed of other manipulation types and authentic news samples. Subsequently, the model is subjected to direct zero-shot inference on a test set constructed exclusively from the held-out target manipulation type and authentic samples. As shown in Table~\ref{fig:crotype}, our methods significantly outperform baselines (e.g., $\text{CORE}_\text{Qwen}$ and $\text{CORE}_\text{Gemma}$ surpass the second-best methods on MDSM by over \textbf{14\%} and \textbf{12\%}, respectively). This superiority validates CORE's design principle of detecting the inherent conflicts essential to fake news. Please refer to the Appendix~\ref{sec:appendix_cross_dataset} for Cross-Dataset Zero-Shot.

\noindent\textbf{Large-scale Training Data.}
To verify CORE's scalability, we evaluate it on the full MDSM and SAMM datasets, together with 200k randomly sampled NewsClippings examples.

As shown in Table~\ref{tab:large-scale}, CORE remains highly effective under large-scale manipulation data of the same type. Across the three benchmarks, $\text{CORE}_\text{Qwen}$ and $\text{CORE}_\text{Gemma}$ outperform the second-best method by an average of \textbf{3.79\%} and \textbf{4.16\%}, respectively, demonstrating strong stability and robustness.

\noindent\textbf{Further Discussion.}
Appendix~\ref{sec:time_sensitive} reports experiments on time-sensitive events beyond the model's pre-training scope, while Appendix~\ref{sec:prompt_vs_training} compares prompting strategies with \text{CORE} training.

\begin{table*}[t]
    \centering
    \caption{\textbf{Loss component ablation} in MBPT and CPT stages on $\text{CORE}_\text{Qwen}$.}
    \label{tab:loss_component_ablation}
    \setlength{\tabcolsep}{4pt}
    \subfloat[Ablation on MBPT loss.
    \label{tab:mbpt_loss_ablation}
    ]{
        \begin{minipage}[t]{0.48\linewidth}
        \centering
        \renewcommand{\arraystretch}{1.1}
        \footnotesize
        \resizebox{\linewidth}{!}{
        \begin{tabular}{l cccc}
            \toprule
            \textbf{Loss Combination} & \textbf{DGM$^4$} & \textbf{MDSM} & \textbf{MMFakeBench} & \textbf{NewsCLIPpings} \\
            \midrule
            w/o $\mathcal{L}_{cl}$ & 61.01 & 69.33 & 75.52 & 66.03 \\
            w/o $\mathcal{L}_{o2vqa}$ & 63.93 & 73.12 & 77.99 & 69.12 \\
            \rowcolor{LightGreen}
            \textbf{Full} ($\mathcal{L}_{cl} + \mathcal{L}_{o2vqa}$) & \textbf{65.18} & \textbf{74.13} & \textbf{79.44} & \textbf{70.88} \\
            \bottomrule
        \end{tabular}
        }
        \end{minipage}
    }
    \hfill
    \subfloat[Ablation on CPT loss.
    \label{tab:cpt_loss_ablation}
    ]{
        \begin{minipage}[t]{0.48\linewidth}
        \centering
        \renewcommand{\arraystretch}{1.1}
        \footnotesize
        \resizebox{\linewidth}{!}{
        \begin{tabular}{l cccc}
            \toprule
            \textbf{Loss Combination} & \textbf{DGM$^4$} & \textbf{MDSM} & \textbf{MMFakeBench} & \textbf{NewsCLIPpings} \\
            \midrule
            w/o $\mathcal{L}_{cacl}$ & 55.17 & 59.30 & 68.94 & 59.22 \\
            w/o $\mathcal{L}_{cr}$ & 63.77 & 73.22 & 78.10 & 69.91 \\
            \rowcolor{LightGreen}
            \textbf{Full} ($\mathcal{L}_{cacl} + \mathcal{L}_{cr}$) & \textbf{65.18} & \textbf{74.13} & \textbf{79.44} & \textbf{70.88} \\
            \bottomrule
        \end{tabular}
        }
        \end{minipage}
    }
    \vspace{-0.2cm}
\end{table*}

\subsection{Ablation Study}
To validate the effectiveness of each component within the CORE framework, 
We designed a series of ablation experiments on $\text{CORE}_\text{Qwen}$:

\noindent\textbf{Data Scale in MBPT.} We evaluated the impact of MBPT data volume (0--75k) on random subsets (500 (300) samples). As shown in Table~\ref{tab:mbpt_data}, insufficient data (0k, 25k) leads to an under-trained Cross-modal Aligner, hindering conflict localization and causing a \textbf{4.47\%} average drop (0k vs. 50k). Conversely, the negligible gap between 50k and 75k indicates performance saturation, confirming that 50k samples suffice for robust alignment.


\noindent\textbf{Loss Discussion in MBPT.} To validate our loss design, we compared it with an MSE-based alternative where ROIAlign~\cite{rcnn} aligns positive visual features $PS$ directly with text features $v_{c_i}$. As shown in Table~\ref{tab:mbpt_loss}, this explicit supervision lags behind contrastive learning~\cite{moco} by \textbf{2.67\%}. This decline stems from disrupting the training consistency with CPT and ROIAlign's limitation in capturing global semantics. For instance, resolving conflicts (e.g., linking ``Ballon d'Or'' to a ``football field'') requires global context that ROIAlign's strict localization severs, thereby degrading performance.

\noindent\textbf{Data Scale in CPT.}
To explore the impact of data scale in CPT, we adjusted the CAC volume from 0k to 38k compared to the standard setting (14k). The results in Table~\ref{tab:cpt_data} reveal that removing the CPT stage (0k) leads to a sharp performance deterioration (avg. \textbf{4.58\%}), underscoring the necessity of CPT. Increasing the volume to 7k improves accuracy but still lags behind 14k (avg. \textbf{0.79\%} gap), suggesting insufficient data for learning semantic conflicts. Conversely, expanding to 38k slightly degrades performance (avg. \textbf{0.43\%}), potentially due to overfitting on manipulation artifacts specific to the SAMM dataset.

\noindent\textbf{Loss Components in MBPT and CPT.}
To further disentangle the contribution of each objective, we remove one loss term in MBPT and CPT, as shown in Table~\ref{tab:loss_component_ablation}. In MBPT, removing $\mathcal{L}_{cl}$ causes a larger average drop than removing $\mathcal{L}_{o2vqa}$ (\textbf{4.44\%} vs. \textbf{1.37\%}), indicating that contrastive alignment provides the primary signal for bridging textual concepts and visual evidence, while the VQA-style objective further stabilizes multimodal instruction learning. In CPT, removing $\mathcal{L}_{cacl}$ leads to the most severe degradation (avg. \textbf{11.75\%}), confirming that explicitly separating conflicting factors is central to conflict perception. Removing $\mathcal{L}_{cr}$ also reduces performance (avg. \textbf{1.16\%}), showing that conflict reasoning supervision helps preserve the model's ability to express and use the learned conflict cues during prediction.

\section{Conclusion}
This paper introduces CORE, a conflict-oriented reasoning framework that enhances MLLMs with explicit conflict-capturing capability for robust and generalizable misinformation detection. By leveraging the newly constructed CAC and a conflict-aware training paradigm, CORE effectively conducts a conflict-perception training and enables rapid adaptation to unseen manipulation types.

\section*{Impact Statement}
\textbf{Positive Impacts:} By focusing on fundamental inconsistencies rather than specific manipulation patterns, CORE improves the robustness of information ecosystems against evolving generative AI. Its high-efficiency reasoning reduces the energy-intensive requirement for frequent full-parameter retraining, supporting sustainable AI deployment.

\textbf{Ethical Considerations \& Mitigation:} 
To prevent the potential {misuse} of our model or the Conflict Attribution Corpus (CAC) for generating more deceptive content, we will implement a {strict data access protocol}. The CAC will be released exclusively for {pure research purposes} under a specialized license that prohibits its use in generative tasks. Furthermore,  all data in the corpus are sourced from public domains to safeguard privacy.

\section*{Acknowledgements}
This work was funded by the National Natural Science Foundation of China (No. 62572166, 62302140, 62502144, 62502142, 62573399) and the Natural Science Foundation of Anhui Province (No. 2508085QF226). The computation is completed on the HPC Platform of Hefei University of Technology.

\bibliography{example_paper}

@String(PAMI = {IEEE Trans. Pattern Anal. Mach. Intell.})

@String(CVPR= {IEEE Conf. Comput. Vis. Pattern Recog.})

@String(ICCV= {Int. Conf. Comput. Vis.})

@String(ECCV= {Eur. Conf. Comput. Vis.})

@String(NIPS= {Adv. Neural Inform. Process. Syst.})

@String(ACMMM= {ACM MM})

@String(ICLR = {Int. Conf. Learn. Represent.})

@String(ICML = {ICML})

@String(NIPS = {Neurips})

@String(EMNLP  = {EMNLP})

@String(PAMI  = {IEEE TPAMI})

@String(CVPR  = {CVPR})

@String(ICCV  = {ICCV})

@String(ECCV  = {ECCV})

@String(NIPS  = {NeurIPS})

@String(ACMMM = {ACM MM})

@String(ICLR  = {ICLR})

@String(EMNLP = {EMNLP})

@String(ACL = {ACL})

@inproceedings{intro1,
  author       = {Kilichbek Haydarov and
                  Aashiq Muhamed and
                  Xiaoqian Shen and
                  Jovana Lazarevic and
                  Ivan Skorokhodov and
                  Chamuditha Jayanga Galappaththige and
                  Mohamed Elhoseiny},
  title        = {Adversarial Text to Continuous Image Generation},
  booktitle    = CVPR,
  pages        = {6316--6326},
  year         = {2024},
}

@inproceedings{intro2,
  author       = {Zhengqi Li and
                  Richard Tucker and
                  Noah Snavely and
                  Aleksander Holynski},
  title        = {Generative Image Dynamics},
  booktitle    = CVPR,
  pages        = {24142--24153},
  year         = {2024},
}

@inproceedings{intro3,
  author       = {Sahar Abdelnabi and
                  Rakibul Hasan and
                  Mario Fritz},
  title        = {Open-Domain, Content-based, Multi-modal Fact-checking of Out-of-Context Images via Online Resources},
  booktitle    = CVPR,
  pages        = {14920--14929},
  year         = {2022},
}

@inproceedings{intro4,
  author       = {Zeyu Lu and
                  Di Huang and
                  Lei Bai and
                  Jingjing Qu and
                  Chengyue Wu and
                  Xihui Liu and
                  Wanli Ouyang},
  editor       = {Alice Oh and
                  Tristan Naumann and
                  Amir Globerson and
                  Kate Saenko and
                  Moritz Hardt and
                  Sergey Levine},
  title        = {Seeing is not always believing: Benchmarking Human and Model Perception
                  of AI-Generated Images},
  booktitle    = NIPS,
  year         = {2023},
}

@InProceedings{intro05,
    author    = {Fanghua Yu and Jinjin Gu and Zheyuan Li and Jinfan Hu and Xiangtao Kong and Xintao Wang and Jingwen He and Yu Qiao and Chao Dong},
    title     = {Scaling Up to Excellence: Practicing Model Scaling for Photo-Realistic Image Restoration In the Wild},
    booktitle = CVPR,
    year      = {2024},
    pages     = {25669-25680}
}

@InProceedings{intro06,
    author    = {Kilichbek Haydarov and Aashiq Muhamed and Xiaoqian Shen and Jovana Lazarevic and Ivan Skorokhodov and Chamuditha Jayanga Galappaththige and Mohamed Elhoseiny},
    title     = {Adversarial Text to Continuous Image Generation},
    booktitle = CVPR,
    year      = {2024},
    pages     = {6316-6326}
}

@InProceedings{intro07,
author = {Liming Jiang and Ren Li and Wayne Wu and Chen Qian and Chen Change Loy},
title = {DeeperForensics-1.0: A Large-Scale Dataset for Real-World Face Forgery Detection},
booktitle = CVPR,
year = {2020},
pages = {2889-2898}
}

@InProceedings{intro08,
author = {Yuezun Li and Xin Yang and Pu Sun and Honggang Qi and Siwei Lyu},
title = {Celeb-DF: A Large-scale Challenging Dataset for DeepFake Forensics},
booktitle = CVPR,
year = {2020},
pages = {3207-3216}
}

@inproceedings{dgm4,
  author       = {Rui Shao and
                  Tianxing Wu and
                  Ziwei Liu},
  title        = {Detecting and Grounding Multi-Modal Media Manipulation},
  booktitle    = CVPR,
  pages        = {6904--6913},
  publisher    = {{IEEE}},
  year         = {2023},
}

@inproceedings{mmfakebench,
  author       = {Xuannan Liu and
                  Zekun Li and
                  Pei{-}Pei Li and
                  Huaibo Huang and
                  Shuhan Xia and
                  Xing Cui and
                  Linzhi Huang and
                  Weihong Deng and
                  Zhaofeng He},
  title        = {MMFakeBench: {A} Mixed-Source Multimodal Misinformation Detection
                  Benchmark for LVLMs},
  booktitle    = ICLR,
  year         = {2025},
}

@inproceedings{samm,
  title = {Beyond Artificial Misalignment: Detecting and Grounding Semantic-Coordinated Multimodal Manipulations},
  author = {Shen, Jinjie and Wang, Yaxiong and Pu, Nan and Cheng, Lechao and Zhong, Zhun},
  booktitle = ACMMM,
  year = {2025},
}

@article{mdsm,
  author       = {Yuchen Zhang and
                  Yaxiong Wang and
                  Yujiao Wu and
                  Lianwei Wu and
                  Li Zhu},
  title        = {The Coherence Trap: When MLLM-Crafted Narratives Exploit Manipulated
                  Visual Contexts},
  journal      = {CoRR},
  year         = {2025},
}

@inproceedings{intro04,
  author       = {Liming Jiang and
                  Ren Li and
                  Wayne Wu and
                  Chen Qian and
                  Chen Change Loy},
  title        = {DeeperForensics-1.0: {A} Large-Scale Dataset for Real-World Face Forgery
                  Detection},
  booktitle    = CVPR,
  pages        = {2886--2895},
  year         = {2020},
}

@inproceedings{intro5,
  author       = {Yuezun Li and
                  Xin Yang and
                  Pu Sun and
                  Honggang Qi and
                  Siwei Lyu},
  title        = {Celeb-DF: {A} Large-Scale Challenging Dataset for DeepFake Forensics},
  booktitle    = CVPR,
  pages        = {3204--3213},
  year         = {2020},
}

@inproceedings{intro6,
  author       = {Rui Shao and
                  Tianxing Wu and
                  Ziwei Liu},
  title        = {Detecting and Recovering Sequential DeepFake Manipulation},
  booktitle    = ECCV,
  pages        = {712--728},
  year         = {2022},
}

@inproceedings{fka-owl,
  author       = {Xuannan Liu and
                  Peipei Li and
                  Huaibo Huang and
                  Zekun Li and
                  Xing Cui and
                  Jiahao Liang and
                  Lixiong Qin and
                  Weihong Deng and
                  Zhaofeng He},
  title        = {FKA-Owl: Advancing Multimodal Fake News Detection through Knowledge-Augmented
                  LVLMs},
  booktitle    = ACMMM,
  pages        = {10154--10163},
  year         = {2024},
}

@article{clc,
  author       = {Yijun Bei and
                  Hengrui Lou and
                  Jinsong Geng and
                  Erteng Liu and
                  Lechao Cheng and
                  Jie Song and
                  Mingli Song and
                  Zunlei Feng},
  title        = {A Large-scale Universal Evaluation Benchmark For Face Forgery Detection},
  journal      = {CoRR},
  volume       = {abs/2406.09181},
  year         = {2024},
  url          = {https://doi.org/10.48550/arXiv.2406.09181},
  doi          = {10.48550/ARXIV.2406.09181},
  eprinttype    = {arXiv},
  eprint       = {2406.09181},
  bibsource    = {dblp computer science bibliography, https://dblp.org}
}

@article{hammer++,
  author       = {Rui Shao and
                  Tianxing Wu and
                  Jianlong Wu and
                  Liqiang Nie and
                  Ziwei Liu},
  title        = {Detecting and Grounding Multi-Modal Media Manipulation and Beyond},
  journal      = pami,
  volume       = {46},
  number       = {8},
  pages        = {5556--5574},
  year         = {2024},
}

@article{qwen2.5vl,
  author       = {Shuai Bai and
                  Keqin Chen and
                  Xuejing Liu and
                  Jialin Wang and
                  Wenbin Ge and
                  Sibo Song and
                  Kai Dang and
                  Peng Wang and
                  Shijie Wang and
                  Jun Tang and
                  Humen Zhong and
                  Yuanzhi Zhu and
                  Ming{-}Hsuan Yang and
                  Zhaohai Li and
                  Jianqiang Wan and
                  Pengfei Wang and
                  Wei Ding and
                  Zheren Fu and
                  Yiheng Xu and
                  Jiabo Ye and
                  Xi Zhang and
                  Tianbao Xie and
                  Zesen Cheng and
                  Hang Zhang and
                  Zhibo Yang and
                  Haiyang Xu and
                  Junyang Lin},
  title        = {Qwen2.5-VL Technical Report},
  journal      = {CoRR},
  year         = {2025},
}

@article{gemma3,
  author       = {Gemma Team},
  title        = {Gemma 3 Technical Report},
  journal      = {CoRR},
  year         = {2025},
}

@article{llama3.2,
  author       = {Llama Team},
  title        = {The Llama 3 Herd of Models},
  journal      = {CoRR},
  year         = {2024},
}

@article{seedvl,
  author       = {Dong Guo and
                  Faming Wu and
                  Feida Zhu and
                  Fuxing Leng and
                  Guang Shi and
                  Haobin Chen and
                  Haoqi Fan and
                  Jian Wang and
                  Jianyu Jiang and
                  Jiawei Wang and
                  Jingji Chen and
                  Jingjia Huang and
                  Kang Lei and
                  Liping Yuan and
                  Lishu Luo and
                  Pengfei Liu and
                  Qinghao Ye and
                  Rui Qian and
                  Shen Yan and
                  Shixiong Zhao and
                  Shuai Peng and
                  Shuangye Li and
                  Sihang Yuan and
                  Sijin Wu and
                  Tianheng Cheng and
                  Weiwei Liu and
                  Wenqian Wang and
                  Xianhan Zeng and
                  Xiao Liu and
                  Xiaobo Qin and
                  Xiaohan Ding and
                  Xiaojun Xiao and
                  Xiaoying Zhang and
                  Xuanwei Zhang and
                  Xuehan Xiong and
                  Yanghua Peng and
                  Yangrui Chen and
                  Yanwei Li and
                  Yanxu Hu and
                  Yi Lin and
                  Yiyuan Hu and
                  Yiyuan Zhang and
                  Youbin Wu and
                  Yu Li and
                  Yudong Liu and
                  Yue Ling and
                  Yujia Qin and
                  Zanbo Wang and
                  Zhiwu He and
                  Aoxue Zhang and
                  Bairen Yi and
                  Bencheng Liao and
                  Can Huang and
                  Can Zhang and
                  Chaorui Deng and
                  Chaoyi Deng and
                  Cheng Lin and
                  Cheng Yuan and
                  Chenggang Li and
                  Chenhui Gou and
                  Chenwei Lou and
                  Chengzhi Wei and
                  Chundian Liu and
                  Chunyuan Li and
                  Deyao Zhu and
                  Donghong Zhong and
                  Feng Li and
                  Feng Zhang and
                  Gang Wu and
                  Guodong Li and
                  Guohong Xiao and
                  Haibin Lin and
                  Haihua Yang and
                  Haoming Wang and
                  Heng Ji and
                  Hongxiang Hao and
                  Hui Shen and
                  Huixia Li and
                  Jiahao Li and
                  Jialong Wu and
                  Jianhua Zhu and
                  Jianpeng Jiao and
                  Jiashi Feng and
                  Jiaze Chen and
                  Jianhui Duan and
                  Jihao Liu and
                  Jin Zeng and
                  Jingqun Tang and
                  Jingyu Sun and
                  Joya Chen and
                  Jun Long and
                  Junda Feng and
                  Junfeng Zhan and
                  Junjie Fang and
                  Junting Lu and
                  Kai Hua and
                  Kai Liu and
                  Kai Shen and
                  Kaiyuan Zhang and
                  Ke Shen},
  title        = {Seed1.5-VL Technical Report},
  journal      = {CoRR},
  year         = {2025},
}

@inproceedings{asap,
  title     = {{ASAP:} Advancing Semantic Alignment Promotes Multi-Modal Manipulation Detecting and Grounding},
  author    = {Zhenxing Zhang and Yaxiong Wang and Lechao Cheng and Zhun Zhong and Dan Guo and Meng Wang},
  booktitle = CVPR ,
  pages        = {4005--4014},
  year      = {2025},
}

@inproceedings{sniffer,
  author       = {Peng Qi and
                  Zehong Yan and
                  Wynne Hsu and
                  Mong{-}Li Lee},
  title        = {Sniffer: Multimodal Large Language Model for Explainable Out-of-Context
                  Misinformation Detection},
  booktitle    = CVPR,
  pages        = {13052--13062},
  year         = {2024},
}

@inproceedings{clip,
  author       = {Alec Radford and
                  Jong Wook Kim and
                  Chris Hallacy and
                  Aditya Ramesh and
                  Gabriel Goh and
                  Sandhini Agarwal and
                  Girish Sastry and
                  Amanda Askell and
                  Pamela Mishkin and
                  Jack Clark and
                  Gretchen Krueger and
                  Ilya Sutskever},
  title        = {Learning Transferable Visual Models From Natural Language Supervision},
  booktitle    = ICML,
  pages        = {8748--8763},
  year         = {2021},
}

@inproceedings{albef,
  author       = {Junnan Li and
                  Ramprasaath R. Selvaraju and
                  Akhilesh Gotmare and
                  Shafiq R. Joty and
                  Caiming Xiong and
                  Steven Chu{-}Hong Hoi},
  title        = {Align before Fuse: Vision and Language Representation Learning with
                  Momentum Distillation},
  booktitle    = NIPS,
  pages        = {9694--9705},
  year         = {2021},
}

@article{tsne,
    author = {Laurens {Van der Maaten} and Geoffrey Hinton},
    title = {Visualizing Data using t-SNE},
    journal = {Journal of Machine Learning Research},
    volume = {9},
    number = {11},
    pages = {2579--2605},
    year = {2008},
}

@misc{googleapi,
  title = {Google Search {API}},
  author = {{Google}},
  year = {2025},
  url = {https://developers.google.com/custom-search/v1/overview}, 
}

@article{gpt4o,
  author       = {GPT-4o Team},
  title        = {GPT-4o System Card},
  journal      = {CoRR},
  year         = {2024},
}

@article{qwen3vlplus,
  author       = {Jinze Bai and
                  Shuai Bai and
                  Shusheng Yang and
                  Shijie Wang and
                  Sinan Tan and
                  Peng Wang and
                  Junyang Lin and
                  Chang Zhou and
                  Jingren Zhou},
  title        = {Qwen-VL: {A} Frontier Large Vision-Language Model with Versatile Abilities},
  journal      = {CoRR},
  year         = {2023},
}

@article{gemini,
  author       = {Gemini Team},
  title        = {Gemini 2.5: Pushing the Frontier with Advanced Reasoning, Multimodality,
                  Long Context, and Next Generation Agentic Capabilities},
  journal      = {CoRR},
  year         = {2025},
}

@inproceedings{finehard,
  author       = {Chunyu Xie and
                  Bin Wang and
                  Fanjing Kong and
                  Jincheng Li and
                  Dawei Liang and
                  Gengshen Zhang and
                  Dawei Leng and
                  Yuhui Yin},
  title        = {{FG-CLIP:} Fine-Grained Visual and Textual Alignment},
  booktitle    = ICML,
  year         = {2025},
}

@inproceedings{transformer,
  author       = {Ashish Vaswani and
                  Noam Shazeer and
                  Niki Parmar and
                  Jakob Uszkoreit and
                  Llion Jones and
                  Aidan N. Gomez and
                  Lukasz Kaiser and
                  Illia Polosukhin},
  title        = {Attention is All you Need},
  booktitle    = NIPS,
  pages        = {5998--6008},
  year         = {2017},
}

@inproceedings{siglip,
  author       = {Xiaohua Zhai and
                  Basil Mustafa and
                  Alexander Kolesnikov and
                  Lucas Beyer},
  title        = {Sigmoid Loss for Language Image Pre-Training},
  booktitle    = ICCV,
  pages        = {11941--11952},
  year         = {2023},
}

@inproceedings{vqa,
  author       = {Stanislaw Antol and
                  Aishwarya Agrawal and
                  Jiasen Lu and
                  Margaret Mitchell and
                  Dhruv Batra and
                  C. Lawrence Zitnick and
                  Devi Parikh},
  title        = {{VQA:} Visual Question Answering},
  booktitle    = ICCV,
  pages        = {2425--2433},
  year         = {2015},
}

@inproceedings{lora,
  author       = {Edward J. Hu and
                  Yelong Shen and
                  Phillip Wallis and
                  Zeyuan Allen{-}Zhu and
                  Yuanzhi Li and
                  Shean Wang and
                  Lu Wang and
                  Weizhu Chen},
  title        = {LoRA: Low-Rank Adaptation of Large Language Models},
  booktitle    = ICLR,
  year         = {2022},
}

@inproceedings{newsclippings,
  author       = {Grace Luo and
                  Trevor Darrell and
                  Anna Rohrbach},
  title        = {NewsCLIPpings: Automatic Generation of Out-of-Context Multimodal Media},
  booktitle    = EMNLP,
  pages        = {6801--6817},
  year         = {2021},
}

@inproceedings{rcnn,
  author       = {Kaiming He and
                  Georgia Gkioxari and
                  Piotr Doll{\'{a}}r and
                  Ross B. Girshick},
  title        = {Mask {R-CNN}},
  booktitle    = ICCV,
  year         = {2017},
}

@inproceedings{moco,
  author       = {Kaiming He and
                  Haoqi Fan and
                  Yuxin Wu and
                  Saining Xie and
                  Ross B. Girshick},
  title        = {Momentum Contrast for Unsupervised Visual Representation Learning},
  booktitle    = CVPR,
  pages        = {9726--9735},
  year         = {2020},
}

@InProceedings{simswap,
    author    = {Renwang Chen and Xuanhong Chen and Bingbing Ni and Yanhao Ge},
    title     = {SimSwap: An Efficient Framework For High Fidelity Face Swapping},
    booktitle = ACMMM,
    year      = {2020},
    pages     = {2003–2011}
}

@InProceedings{infoswap,
    author    = {Gege Gao and Huaibo Huang and Chaoyou Fu and Zhaoyang Li and Ran He},
    title     = {Information Bottleneck Disentanglement for Identity Swapping},
    booktitle = CVPR,
    year      = {2021},
    pages     = {3404-3413}
}

@InProceedings{hfgi,
    author    = {Tengfei Wang and Yong Zhang and Yanbo Fan and Jue Wang and Qifeng Chen},
    title     = {High-Fidelity GAN Inversion for Image Attribute Editing},
    booktitle = CVPR,
    year      = {2022},
    pages     = {11379-11388}
}

@InProceedings{styleclip,
    author    = {Or Patashnik and Zongze Wu and Eli Shechtman and Daniel Cohen-Or and Dani Lischinski},
    title     = {StyleCLIP: Text-Driven Manipulation of StyleGAN Imagery},
    booktitle = ICCV,
    year      = {2021},
    pages     = {2085-2094}
}

@inproceedings{intro11,
  author       = {Tom B. Brown and
                  Benjamin Mann and
                  Nick Ryder and
                  Melanie Subbiah and
                  Jared Kaplan and
                  Prafulla Dhariwal and
                  Arvind Neelakantan and
                  Pranav Shyam and
                  Girish Sastry and
                  Amanda Askell and
                  Sandhini Agarwal and
                  Ariel Herbert{-}Voss and
                  Gretchen Krueger and
                  Tom Henighan and
                  Rewon Child and
                  Aditya Ramesh and
                  Daniel M. Ziegler and
                  Jeffrey Wu and
                  Clemens Winter and
                  Christopher Hesse and
                  Mark Chen and
                  Eric Sigler and
                  Mateusz Litwin and
                  Scott Gray and
                  Benjamin Chess and
                  Jack Clark and
                  Christopher Berner and
                  Sam McCandlish and
                  Alec Radford and
                  Ilya Sutskever and
                  Dario Amodei},
  title        = {Language Models are Few-Shot Learners},
  booktitle    = NIPS,
  year         = {2020},
}

@inproceedings{intro12,
  author       = {Zhenhailong Wang and
                  Manling Li and
                  Ruochen Xu and
                  Luowei Zhou and
                  Jie Lei and
                  Xudong Lin and
                  Shuohang Wang and
                  Ziyi Yang and
                  Chenguang Zhu and
                  Derek Hoiem and
                  Shih{-}Fu Chang and
                  Mohit Bansal and
                  Heng Ji},
  title        = {Language Models with Image Descriptors are Strong Few-Shot Video-Language
                  Learners},
  booktitle    = NIPS,
  year         = {2022},
}

@inproceedings{intro13,
  author       = {Aman Madaan and
                  Shuyan Zhou and
                  Uri Alon and
                  Yiming Yang and
                  Graham Neubig},
  title        = {Language Models of Code are Few-Shot Commonsense Learners},
  booktitle    = EMNLP,
  pages        = {1384--1403},
  year         = {2022},
}

@inproceedings{yaxiongAIGCDet1,
  author       = {Zhenxing Zhang and
                  Yaxiong Wang and
                  Lechao Cheng and
                  Zhun Zhong and
                  Dan Guo and
                  Meng Wang},
  title        = {{ASAP:} Advancing Semantic Alignment Promotes Multi-Modal Manipulation Detecting and Grounding},
  booktitle    = CVPR,
  pages        = {4005--4014},
  year         = {2025},
}

@inproceedings{yaxiongAIGCDet2,
  title={Cultivating Forensic Reasoning for Generalizable Multimodal Manipulation Detection},
  author={Yuchen Zhang and Yaxiong Wang and Kecheng Han and Yujiao Wu and Lianwei Wu and Li Zhu and Zhedong Zheng},
  booktitle= ACL,
  year={2026}
}

@inproceedings{yaxiongAIGCDet3,
  title={Generating Attribution Reports for Manipulated Facial Images: A Dataset and Baseline},
  author={Jingchun Lian and Lingyu Liu and Yaxiong Wang and Yujiao Wu and Lianwei Wu and Li Zhu and Zhedong Zheng},
  booktitle= ACL,
  year={2026}
}
\bibliographystyle{icml2026}

\newpage
\appendix
\onecolumn
\section{Implementation Details}
\label{appendix-implementation_details}
During MBPT, we optimize the Modality Connector and the LLM with a uniform learning rate of $1 \times 10^{-4}$ for 1 epoch, while the Vision Encoder is frozen. In CPT, we continue to optimize the Modality Connector and the LLM for 3 epochs, while the Vision Encoder remains frozen. In RA, we optimize only the LLM for 3 to 8 epochs, depending on the scale of the training data. All experiments are conducted on devices equipped with 4 NVIDIA H200 GPUs.

\section{Benchmark Brief Introduction}
\label{appendix-bench}
To ensure a rigorous evaluation of generalization, we select multiple representative benchmarks that cover a wide spectrum of manipulation types. The defining features of these datasets are outlined below:

\noindent\textbf{NewsCLIPpings} focuses on the out-of-context (OOC) threat scenario: it provides a dataset where both the image and text are individually unmanipulated, but are automatically mismatched to create semantic or entity inconsistencies.

\noindent\textbf{DGM$^4$} employs a random manipulation pipeline: it randomly manipulates a part of the news's image or caption (e.g., randomly replacing some words in the caption with other words).

\noindent\textbf{MMFakeBench} introduces a comprehensive benchmark for mixed-source MMD: it includes 3 critical sources (textual veracity distortion, visual veracity distortion, and cross-modal consistency distortion) along with 12 sub-categories of misinformation forgery types.

\noindent\textbf{SAMM} pioneers the detection of semantically-coordinated manipulations: it first applies SOTA image manipulations, and then generates contextually-plausible, semantically consistent textual narratives designed to reinforce the visual deception.

\noindent\textbf{MDSM} utilizes an adversarial pipeline that leverages MLLMs to simulate high-risk disinformation: it first alters images using SOTA editing techniques, and then pairs them with MLLM-generated deceptive texts that maintain semantic consistency with the visual manipulations.

\section{Concepts in Section~3}
\begin{figure}[H]
    \centering
    \includegraphics[width=\linewidth]{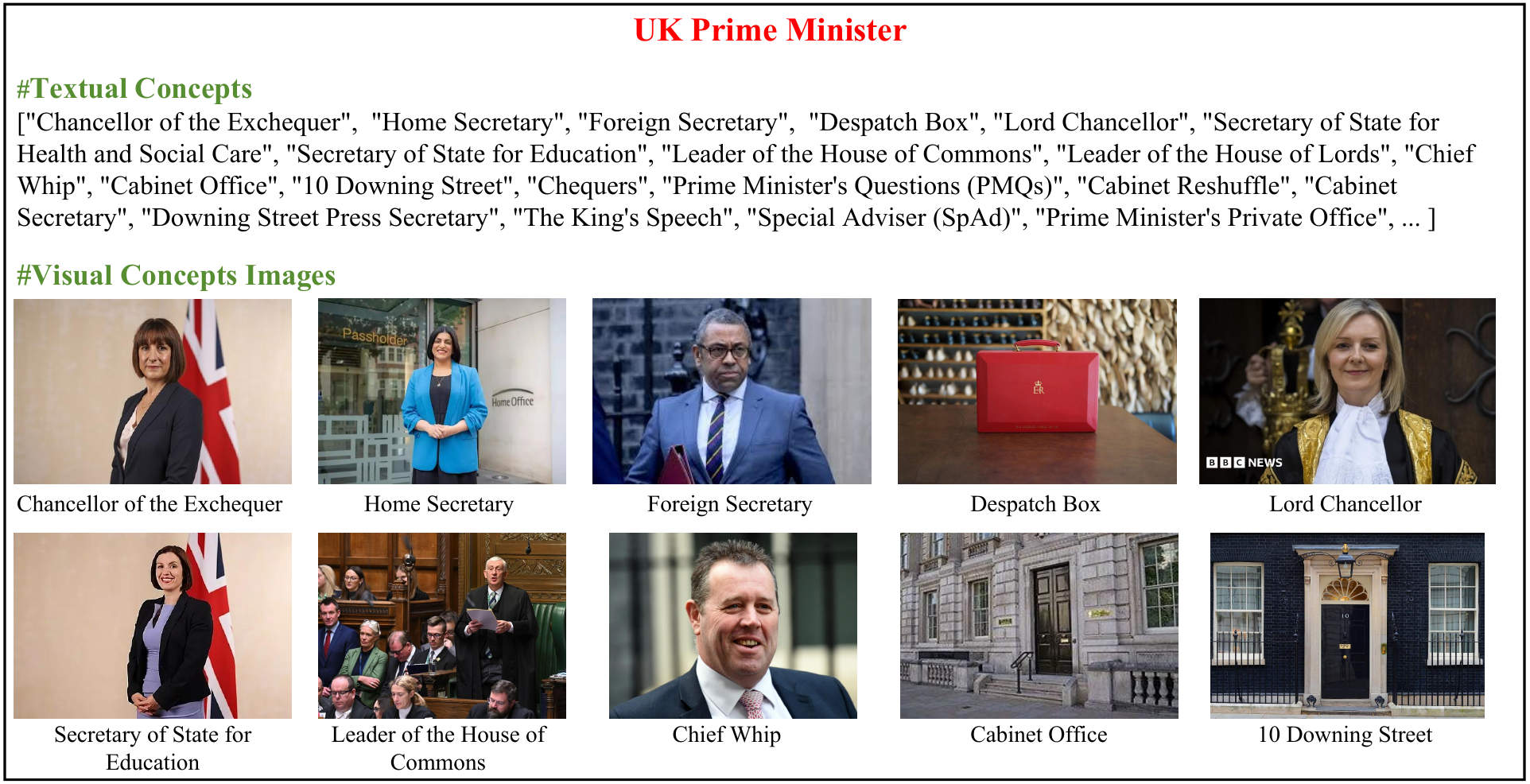}
    \caption{Examples of UK Prime Ministers.}
    \label{fig:primeminister_uk}
\end{figure}

\begin{figure}[H]
    \centering
    \includegraphics[width=1\linewidth]{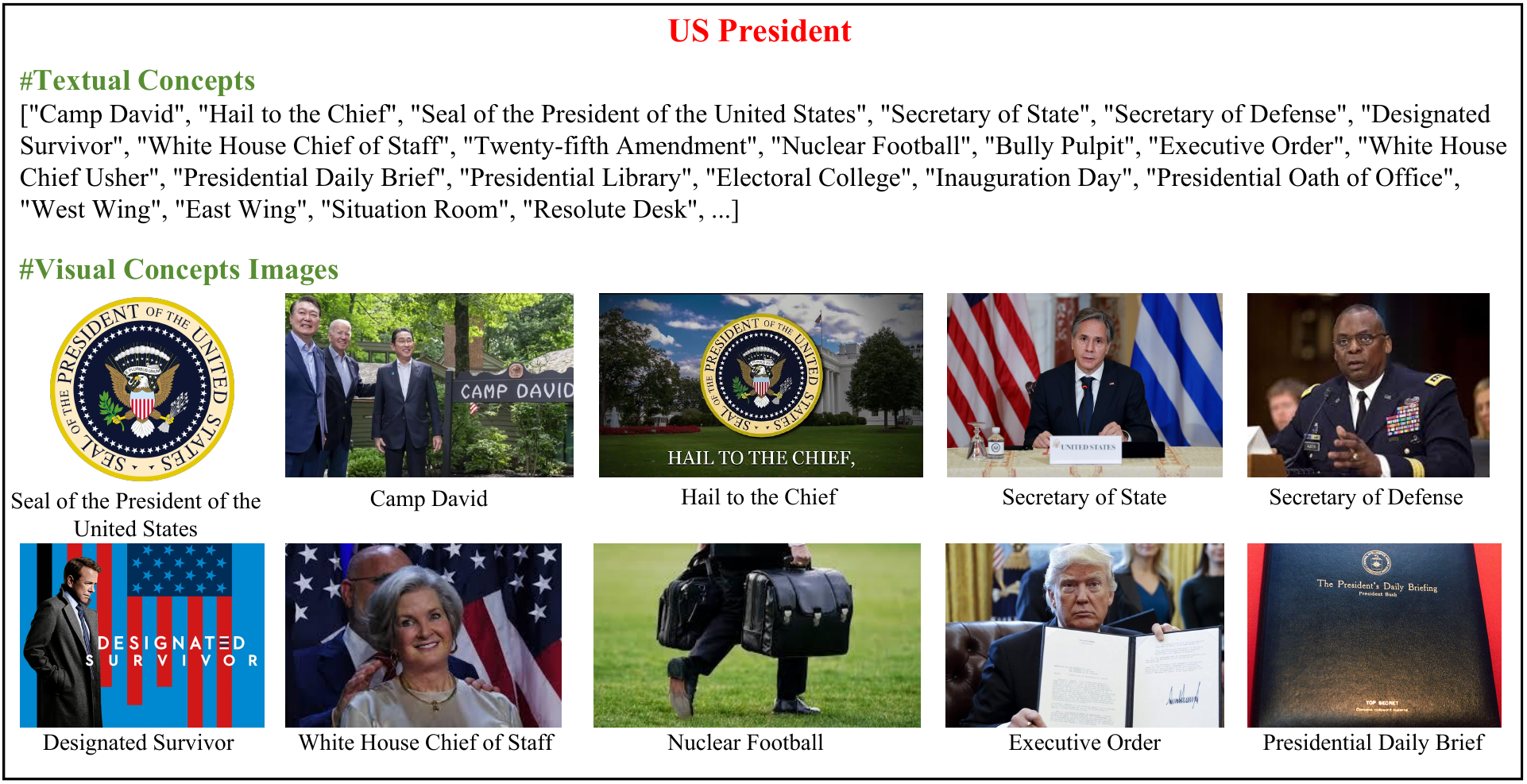}
    \caption{Examples of US President.}
    \label{fig:president_us}
\end{figure}

\begin{figure}[H]
    \centering
    \includegraphics[width=1\linewidth]{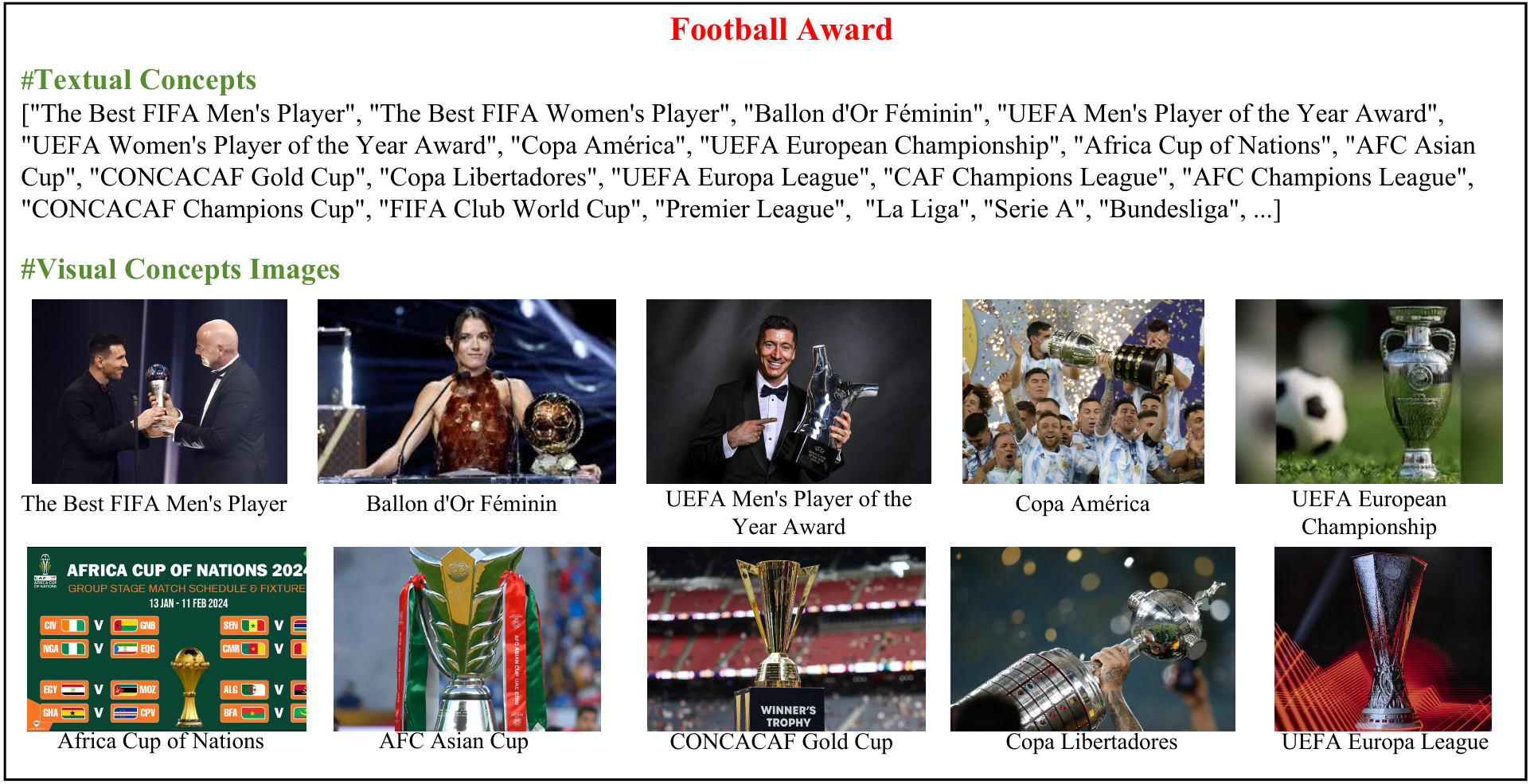}
    \caption{Examples of Football Award.}
    \label{fig:football_award}
\end{figure}

Figure~\ref{fig:primeminister_uk}-\ref{fig:football_award} illustrates some of the textual concepts and images used in our experiments in Section~3.


\section{Examples of CAC}
\begin{figure}[H]
    \centering
    \includegraphics[width=1\linewidth]{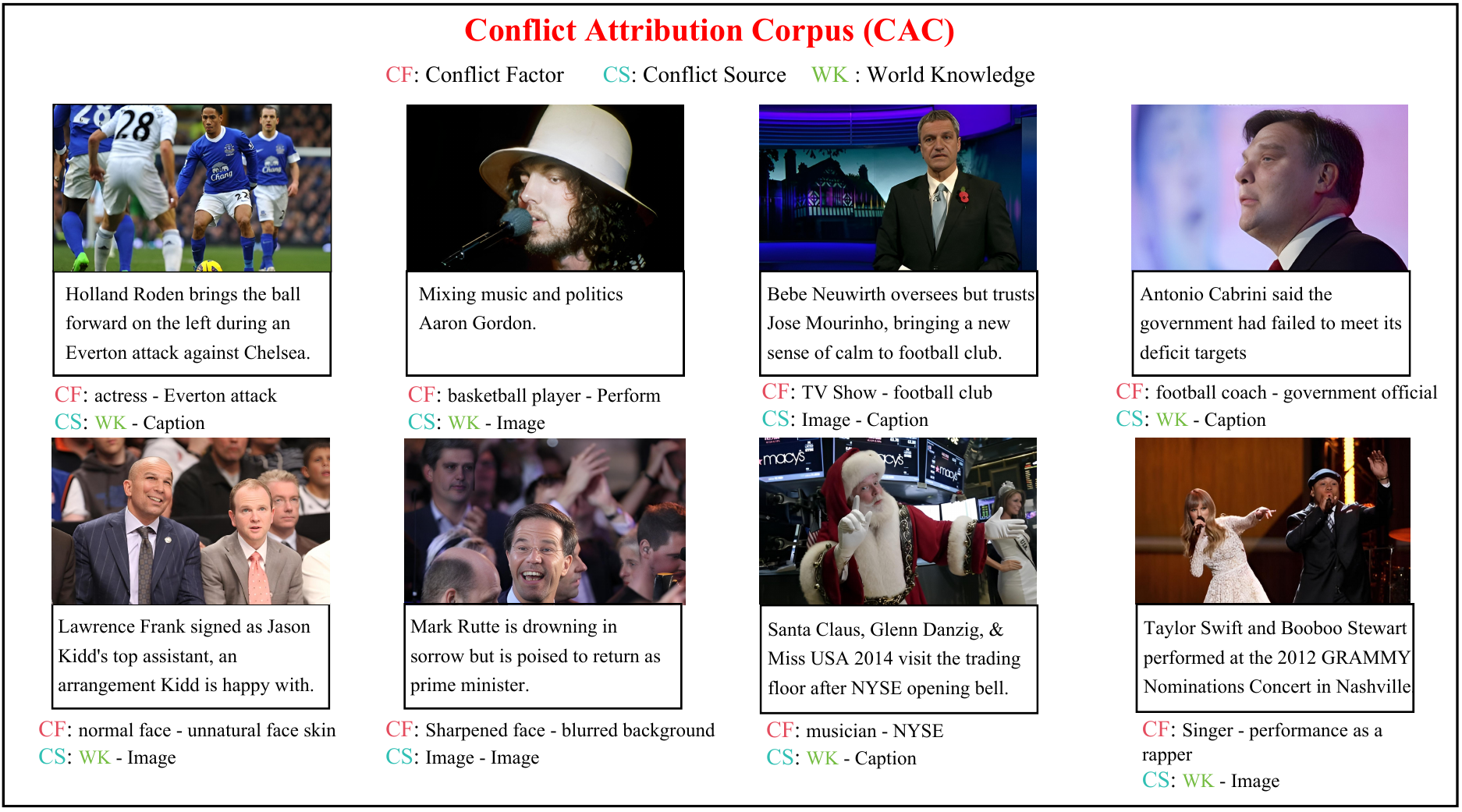}
    \caption{Examples of CAC}
    \label{fig:placeholder_cac}
\end{figure}

\section{Examples of FineHARD}
\begin{figure}[H]
    \centering
    \includegraphics[width=1\linewidth]{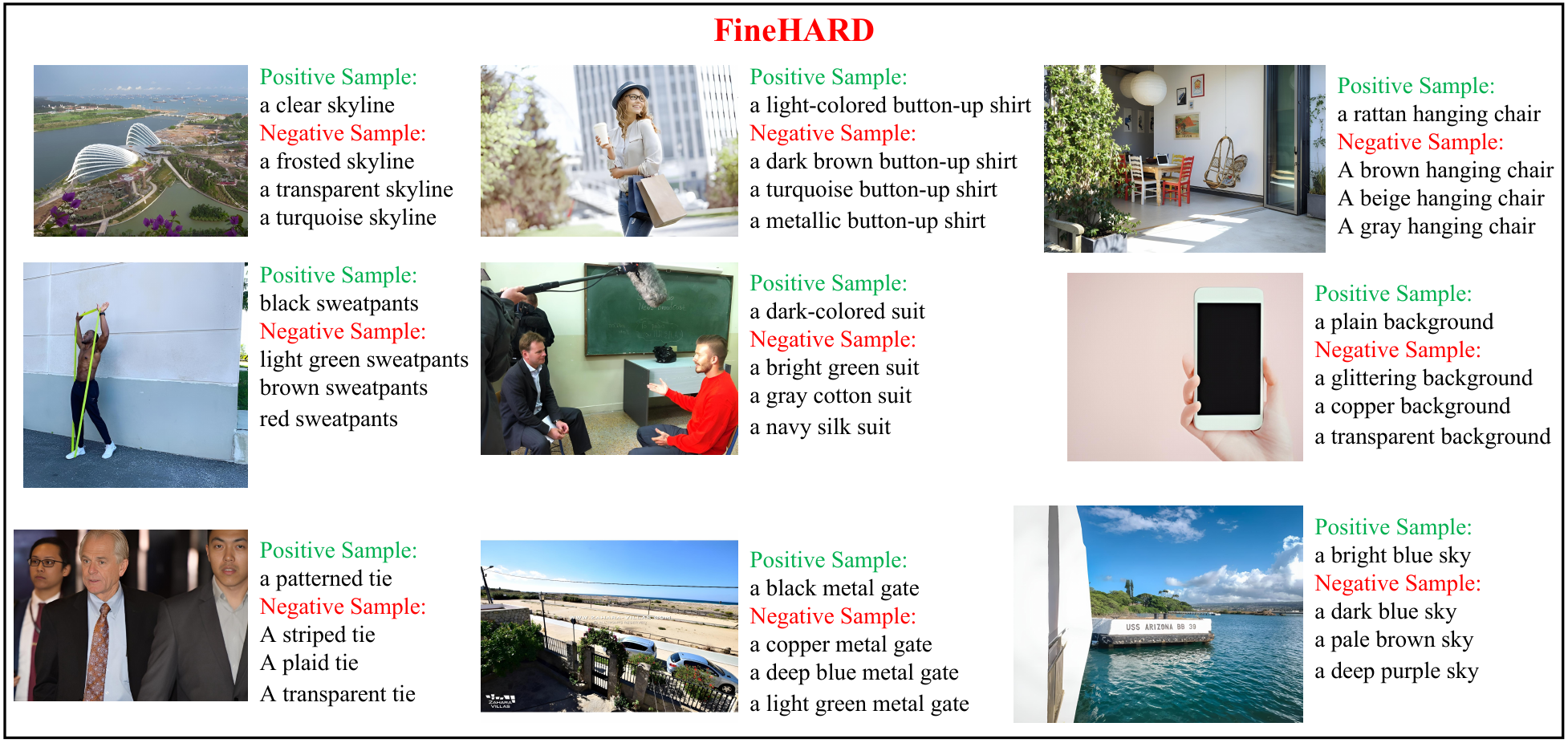}
    \caption{Examples of FineHARD.}
    \label{fig:placeholder_finehard}
\end{figure}

\section{Feature Extraction Details in Section~3}
\label{appendixc-feature_extraction}
For visual features, the image is passed through the vision encoder~$\mathcal{E}_V$ and the modality connector~$\mathcal{P}$ to obtain a sequence of visual feature embeddings. For textual features, the text concept is fed into the LLM, and we extract the corresponding sequence of hidden states from its final layer. Subsequently, we apply an averaging operation to both the visual and textual feature sequences, transforming each into a tensor of shape $[1, \text{hidden size}]$.

\section{Construction of World Knowledge Evaluation Benchmark}
\label{sec:wk_benchmark}

To assess whether models possess the necessary background knowledge for fake news detection, we constructed the World Knowledge Evaluation Benchmark. This benchmark comprises 200 multiple-choice questions meticulously designed to cover a diverse range of domains frequently targeted by misinformation, including current events (e.g., political leaders, cultural awards), social movements, geography, history, and science. 

\begin{figure}[htbp]
    \centering
    \includegraphics[width=1\linewidth]{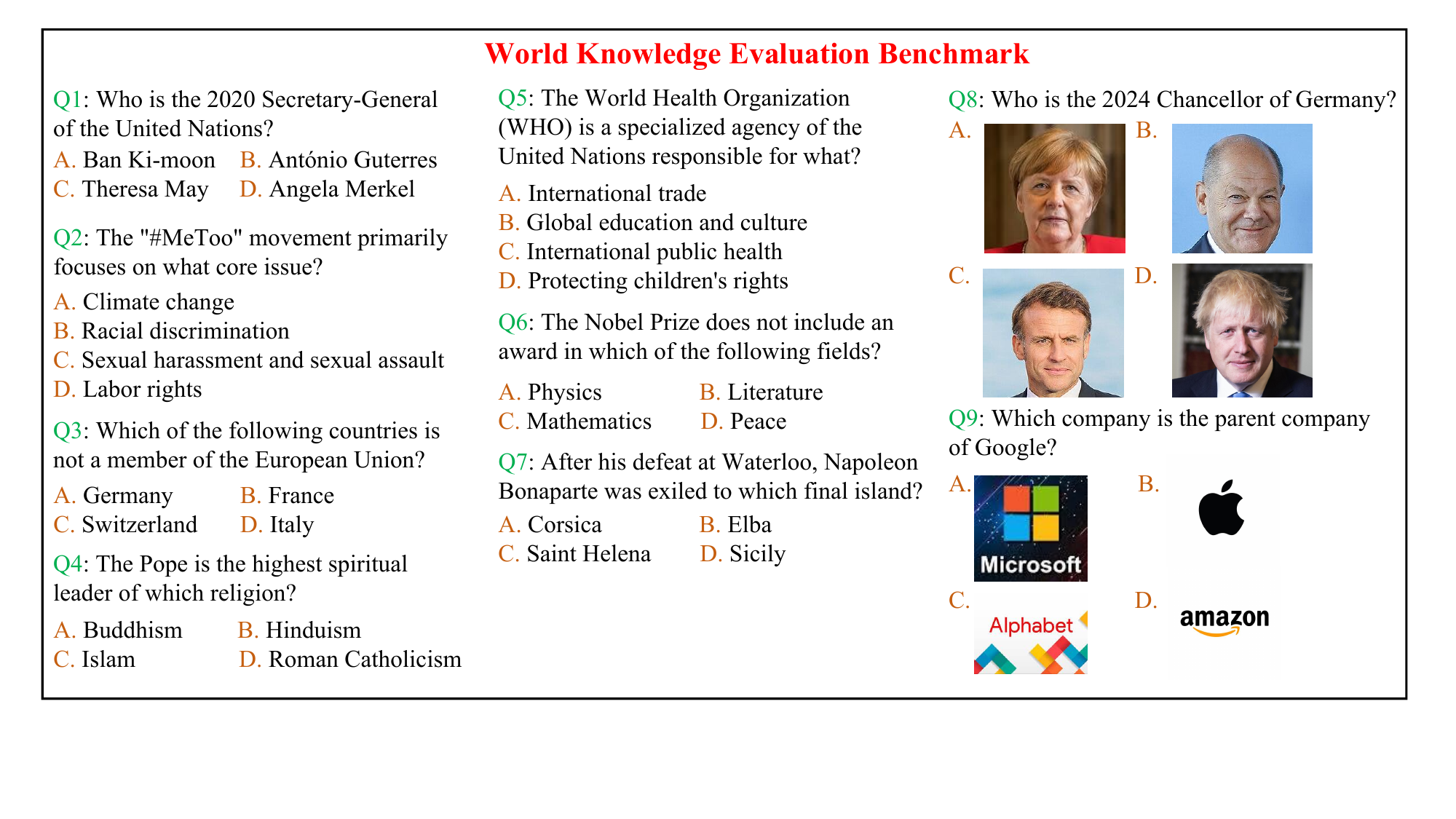}
    \caption{Examples of the World Knowledge Evaluation Benchmark, highlighting the use of semantically plausible distractors.}
    \label{fig:wk_benchmark}
\end{figure}

A key aspect of our construction process was the creation of plausible, semantically-related distractors. Unlike standard QA datasets where incorrect answers are often random, our distractors share the same semantic category as the ground truth. For example, distractors for the question "Which film won the Academy Award for Best Picture in 2024?" include other highly-nominated films from the same ceremony (e.g., "Barbie," "Poor Things"). Similarly, distractors for the 2024 German Chancellor include the former chancellor and other contemporary European leaders. This design ensures the benchmark evaluates precise knowledge rather than coarse categorical association. Figure~\ref{fig:wk_benchmark} illustrates several examples from this benchmark.

To address the need for transparency regarding data provenance and curation, we provide the specific construction details below, focusing on data sources, collection pipelines, and domain-balancing criteria.

\subsection{Data Sources}
To ensure factual accuracy and relevance, we curated data from two primary streams:
\begin{itemize}
    \item \textbf{Authoritative Knowledge Bases:} Static facts (e.g., geography, history, scientific definitions) were sourced from high-reliability encyclopedic sources, primarily Wikipedia and Britannica, ensuring that the ground truth is indisputable.
    \item \textbf{Verified News Outlets:} Dynamic facts involving current events (e.g., 2024 political leaders, recent cultural awards) were cross-referenced against major reputable news agencies (e.g., BBC, Reuters, AP).
\end{itemize}

\subsection{Collection and Generation Pipeline}
The question collection process followed a "Human-Guided, AI-Assisted" paradigm involving three stages:
\begin{enumerate}
    \item \textbf{Entity and Topic Extraction:} We first identified high-frequency entities and topics appearing in the news, such as political figures (e.g., US Presidents, European leaders), celebrities, and major global organizations (e.g., WHO, UN). This ensures the benchmark evaluates knowledge directly relevant to the downstream detection task.
    \item \textbf{Distractor-Aware Question Generation:} As noted in the introduction, we enforced the creation of \textit{semantically plausible distractors} to test specific factual knowledge rather than simple elimination capabilities.
    \item \textbf{Adversarial Filtering:} We employed GPT-4o to review the questions. Any question that could be answered solely through linguistic bias or simple elimination without specific knowledge was discarded or rewritten.
\end{enumerate}

\subsection{Selection and Domain-Balancing Criteria}
To prevent domain bias, we established a strict taxonomy covering five distinct categories. The 200 questions were balanced to ensure broad coverage of the "world knowledge" typically exploited in multimodal misinformation. The domain distribution and selection criteria are defined in Table~\ref{tab:knowledge_domains}.

\begin{table}[H]
\centering
\caption{Domain Distribution and Selection Criteria for the World Knowledge Benchmark.}
\label{tab:knowledge_domains}
\begin{tabular}{l|c|p{8cm}}
\toprule
\textbf{Domain} & \textbf{Ratio} & \textbf{Selection Criteria} \\
\midrule
\textbf{Current Events} & 30\% & Focuses on heads of state (e.g., German Chancellor, UK PM), major elections, and shifting geopolitical alliances active between 2020--2024. \\
\midrule
\textbf{Culture \& Entertainment} & 20\% & Includes major awards (Oscars, Grammys, Ballon d'Or) and celebrity identities often targeted by "face swap" deepfakes. \\
\midrule
\textbf{Social \& Political Movements} & 15\% & Covers defining movements (e.g., \#MeToo, BLM, Climate Change initiatives) to test understanding of complex social contexts. \\
\midrule
\textbf{History \& Geography} & 20\% & Tests static knowledge often used as background context in news (e.g., EU membership status, locations of landmarks, historical exiles). \\
\midrule
\textbf{Science \& Public Health} & 15\% & Focuses on verifiable scientific consensus and major organizations (e.g., WHO mandates, COVID-19 terminology) frequently subject to medical misinformation. \\
\bottomrule
\end{tabular}
\end{table}

\paragraph{Final Human Verification} All questions underwent a final manual verification round by the authors to confirm that: (1) the ground truth is unambiguous, and (2) the distractors are factually incorrect but contextually relevant.

\section{Prompt Templates and Validation Protocols for CAC Construction}
\label{sec:appendix_prompts}

To ensure reproducibility and transparency regarding the construction of the Conflict Attribution Corpus (CAC), we provide the exact prompts used at each stage of the pipeline, along with the specific validation criteria employed by the MLLM expert pool and human annotators. As detailed in Section 4.1, our pipeline utilizes an expert pool $\mathcal{M} = \{\text{GPT-4o}, \text{Gemini-2.5-Pro}, \text{Qwen3-VL-Plus}\}$.

\subsection{Background Knowledge Collection Prompts}
As described in the implementation details, this stage leverages an MLLM to bridge the gap between the raw news sample and external knowledge. The process follows a specific pipeline: analyzing the image and caption to extract key semantic information (e.g., time, event, celebrities), combining this into search queries, conducting separate web and image searches via the Google Search API, and finally validating the relevance of retrieved materials.

\begin{table}[H]
    \centering
    \caption{Prompts for Background Knowledge Collection (Google Search API Stage).}
    \label{tab:prompt_background}
    \begin{tabular}{p{0.95\linewidth}}
        \toprule
        \textbf{Step 1: Semantic Extraction and Query Generation} \\
        \midrule
        \textbf{Input:} Image $I$, Caption $T$, Prior Manipulation Info $P$ (from SAMM). \\
        \textbf{Prompt:} ``You are an investigative expert assisting in fact-checking. 
        1. Analyze the provided news image and caption to extract key semantic information, specifically identifying: \textbf{Time} (dates, eras), \textbf{Events} (political rallies, ceremonies, conflicts), and \textbf{Celebrities/Entities} (politicians, organizations).
        2. Based on these extracted semantics, generate 3 distinct search queries for the Google Search API to gather external evidence. 
        3. Ensure queries are suitable for both \textit{Textual Search} (to find news articles) and \textit{Image Search} (to find original source photos).
        Return the output as a JSON list of query strings.'' \\
        \midrule
        \textbf{Step 2: Retrieved Material Relevance Validation} \\
        \midrule
        \textbf{Input:} Image $I$, Caption $T$, Retrieved Material $R$ (Text Snippet or Image). \\
        \textbf{Prompt:} ``You are a data filter. Review the retrieved material $R$ obtained from Google Search. Does $R$ provide highly relevant factual context, background details, or visual evidence related to the entities and events extracted from the query image $I$ and caption $T$? 
        \textbf{Criterion:} Answer 'YES' only if $R$ is directly relevant and useful for verifying the authenticity of the news; otherwise, answer 'NO'.'' \\
        \bottomrule
    \end{tabular}
\end{table}

\subsection{Conflict Rationale Generation Prompts}
The core logic of CORE relies on identifying \textit{why} a sample is fake. We task a randomly selected MLLM from $\mathcal{M}$ to generate this rationale, which is then cross-validated by the other experts in the pool.

\begin{table}[H]
    \centering
    \caption{Prompts for Conflict Rationale Generation and Validation.}
    \label{tab:prompt_rationale}
    \begin{tabular}{p{0.95\linewidth}}
        \toprule
        \textbf{Generation Prompt} \\
        \midrule
        \textbf{System:} You are a forensics expert specializing in multimodal misinformation. \\
        \textbf{Input:} Image $I$, Caption $T$, Validated Background Knowledge $B$ (from Step 2), Prior Manipulation Info $P$. \\
        \textbf{Instruction:} ``This news item is known to be manipulated based on Prior Info $P$. Using the external Background Knowledge $B$ gathered via search, explain specifically \textit{why} it is false. Focus on the intrinsic conflict between the visual content, the textual caption, and world knowledge. Do not simply state it is fake; describe the logical contradiction in detail.'' \\
        \midrule
        \textbf{Validation Prompt} \\
        \midrule
        \textbf{System:} You are a quality assurance auditor for misinformation detection. \\
        \textbf{Input:} Image $I$, Caption $T$, Validated Background Knowledge $B$, Generated Rationale $R$. \\
        \textbf{Instruction:} ``Evaluate the validity of the provided Rationale $R$ against the evidence. Check two criteria: \\
        1. \textbf{Consistency:} Does $R$ accurately reflect the visible content in $I$ and $T$? \\
        2. \textbf{Factual Support:} Is the logical contradiction described in $R$ fully supported by the Background Knowledge $B$? \\
        Return `Valid' only if both criteria are strictly met; otherwise return `Invalid'.'' \\
        \bottomrule
    \end{tabular}
\end{table}

\subsection{Conflict Structuring Prompts}
This stage distills the natural language rationale into the structured format $<C_1, C_2, S_1, S_2>$ required for the CPT stage.

\begin{table}[H]
    \centering
    \caption{Prompts for Conflict Structuring and Validation.}
    \label{tab:prompt_structuring}
    \begin{tabular}{p{0.95\linewidth}}
        \toprule
        \textbf{Structuring Prompt} \\
        \midrule
        \textbf{Input:} Image $I$, Caption $T$, Approved Rationale $R_{expl}$. \\
        \textbf{Instruction:} ``Based on the provided rationale, extract the two specific conflicting elements (Conflict Factors) and their origins (Conflict Sources). \\
        Definitions: \\
        - \textit{Conflict Factor ($C$)}: The specific concept (e.g., `USA President', `Ballon d'Or'). \\
        - \textit{Conflict Source ($S$)}: Where $C$ comes from. Must be one of [`Image', `Caption', `World Knowledge']. \\
        
        Output strictly in this JSON format: \\
        \{``C1'': ``...'', ``S1'': ``...'', ``C2'': ``...'', ``S2'': ``...''\}'' \\
        \midrule
        \textbf{Validation Prompt} \\
        \midrule
        \textbf{System:} You are a data quality auditor ensuring alignment between textual reasoning and structured data. \\
        \textbf{Input:} Approved Rationale $R_{expl}$, Generated JSON Structure $J$. \\
        \textbf{Instruction:} ``Verify if the JSON object $J$ accurately distills the conflict described in Rationale $R_{expl}$. Check the following: \\
        1. \textbf{Content Match:} Do $C_1$ and $C_2$ represent the exact contradictory concepts mentioned in $R_{expl}$? \\
        2. \textbf{Source Attribution:} Are the sources $S_1$ and $S_2$ correctly identified as `Image', `Caption', or `World Knowledge' according to the rationale? \\
        Return `Valid' only if the extraction is precise and the format is correct; otherwise return `Invalid'.'' \\
        \bottomrule
    \end{tabular}
\end{table}

\subsection{Final Human Verification Protocol}
While the automated pipeline ensures scalability, we introduced a rigorous human-in-the-loop verification step to ensure the quality of the CAC dataset. We employed 5 distinct annotators to validate the dataset.

\textbf{Sampling Strategy:} We randomly select 1k samples, ensuring balanced coverage across the various conflict source distributions.

\textbf{Validation Rubric:} Annotators were instructed to reject or correct samples based on the following criteria:
\begin{enumerate}
    \item \textbf{Conflict Existence:} Does a logical contradiction actually exist between $C_1$ and $C_2$?
    \item \textbf{Source Accuracy:} Are $S_1$ and $S_2$ correctly attributed? (e.g., if $S_1$ is labeled `World Knowledge', does it rely on external facts rather than visual cues?)
    \item \textbf{Granularity Check:} Are the conflict factors fine-grained concepts (e.g., ``red tie'' vs ``blue tie'') rather than abstract descriptions (e.g., ``fake image'' vs ``real text'')?
\end{enumerate}

\textbf{Outcome:} Ultimately, 993 samples passed human verification (a pass rate of 99.3\%), indicating high reliability in the automated generation process.

\section{Evaluation on Time-Sensitive Events}
\label{sec:time_sensitive}

To address the concern regarding the model's applicability when handling \textit{time-sensitive events that fall outside the model’s pre-trained knowledge scope}, we conducted an additional evaluation focusing on emerging misinformation.

\begin{figure}[H]
    \centering
    \includegraphics[width=0.7\linewidth]{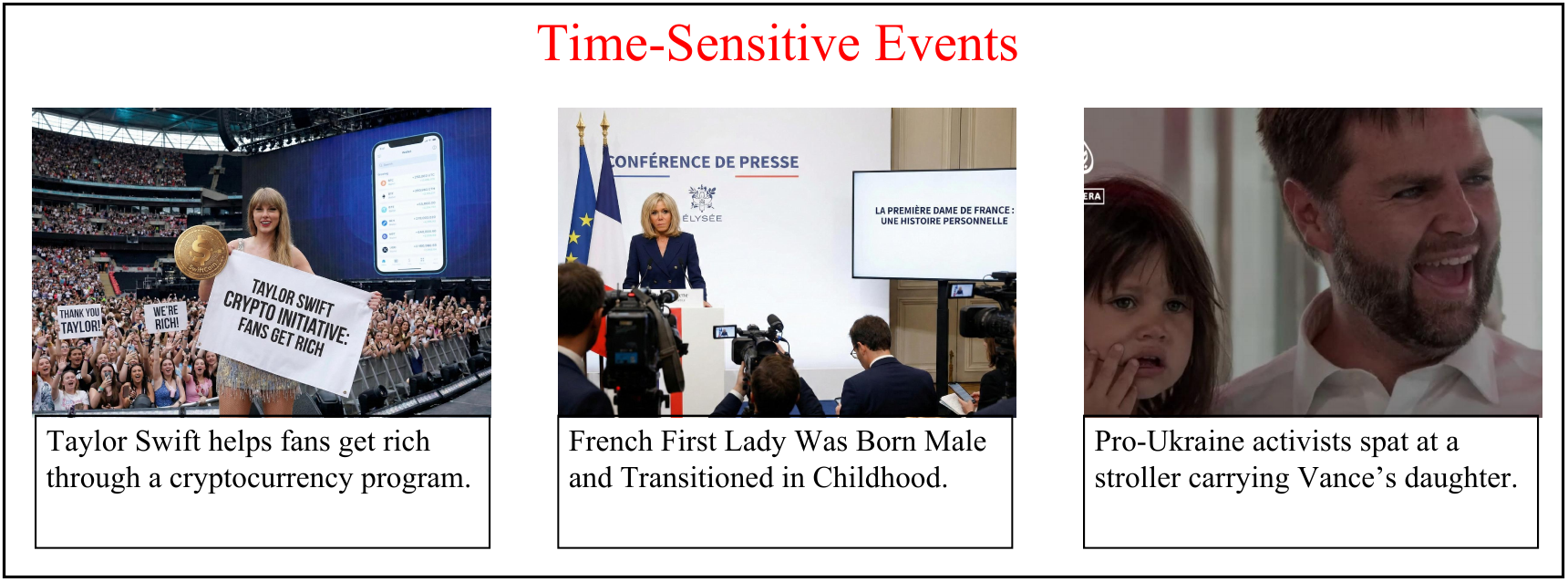}
    \caption{Examples of time-sensitive fake news samples (2025) used in our evaluation. Despite lacking specific pre-trained knowledge of these recent events, CORE successfully identifies the misinformation by detecting intrinsic logical conflicts and cross-modal inconsistencies, whereas baseline models often fail.}
    \label{fig:time_sensitive_examples}
\end{figure}

\textbf{Dataset Construction.} We collected a distinct dataset consisting of 100 high-risk, time-sensitive fake news samples from social media platforms. To rigorously test the "out-of-scope" condition, the majority of these events occurred in 2025, ensuring they post-date the training cut-off of the foundation models and our training corpus. These samples simulate real-world "zero-day" misinformation scenarios where specific world knowledge is absent from the model's parametric memory. Figure~\ref{fig:time_sensitive_examples} visualizes three examples from this collected dataset.

\textbf{Baselines.} We benchmarked $\text{CORE}_\text{Qwen}$ against three representative state-of-the-art methods: HAMMER, AMD and FKA-Owl.

\textbf{Results and Analysis.} The quantitative results are presented in Table~\ref{tab:time_sensitive_results}. While baseline methods struggle significantly due to their reliance on specific patterns or outdated knowledge bases (ranging from 44\% to 52\% accuracy), CORE achieves an accuracy of 74\%. 

This superior performance indicates that while $\text{CORE}_\text{Qwen}$ may lack specific knowledge of the exact 2025 event (e.g., the specific outcome of a new election), its \textit{Conflict-Oriented Reasoning} paradigm allows it to identify falsified news by detecting: (1) \textbf{Intrinsic Logic Violations:} Contradictions within the text or between visual elements that violate general physical or logical rules (which remain constant regardless of the year). (2) \textbf{Cross-Modal Inconsistencies:} Discrepancies between the provided image and the textual claim that do not require specific entity knowledge to detect (e.g., emotional mismatch, scene inconsistency).

\begin{table}[H]
    \centering
    \setlength{\tabcolsep}{10pt}
    \caption{Performance comparison on Time-Sensitive Fake News. This dataset comprises events falling outside the pre-trained knowledge scope of the models.}
    
    \renewcommand{\arraystretch}{1.1}
    \footnotesize
    
    \begin{tabular}{lc}
        \toprule
        \textbf{Method} & \textbf{Accuracy (\%)} \\
        \midrule
        AMD & 44.0 \\
        HAMMER & 48.0 \\
        FKA-Owl & 52.0 \\
        
        \rowcolor{LightGreen}
        \textbf{$\text{CORE}_\text{Qwen}$} & \textbf{74.0} \\
        \bottomrule
    \end{tabular}
    \label{tab:time_sensitive_results}
    \vspace{-0.3cm}
\end{table}

\section{Cross-Dataset Zero-Shot Generalization}
\label{sec:appendix_cross_dataset}

We conduct a rigorous \textbf{Cross-Dataset Generalization} experiment. This setting is designed to evaluate the model's performance in a true open-world scenario, where the testing data originates from a completely different source distribution with potentially unseen manipulation pipelines compared to the training data.

\textbf{Experimental Setup.} We use four benchmarks: NewsCLIPpings, DGM$^4$, MMFakeBench, and MDSM.
Specifically, to evaluate performance on a target dataset (e.g., NewsCLIPpings), we trained the models on a subset composed of the remaining three datasets (e.g., DGM$^4$, MMFakeBench, and MDSM). To simulate a low-resource adaptation scenario effectively, we randomly sampled a total of only 3,000 samples from the union of the three training datasets. The models were then evaluated directly on the full test set of the held-out target dataset in a zero-shot manner.

\textbf{Results and Analysis.} The quantitative results are reported in Table~\ref{tab:cross_dataset_results}.
As observed, existing methods struggle significantly in this cross-dataset setting. For instance, AMD, which relies heavily on specific manipulation traces and prior information, drops to near-random performance (40.2\% - 46.1\%) when the testing distribution shifts. Similarly, FKA-Owl and HAMMER exhibit limited robustness, with accuracies hovering between 43\% and 54\%. This suggests that these baselines tend to overfit to the specific data distributions or manipulation artifacts present in their training domains, failing to generalize to the semantic or physical inconsistencies inherent in unseen datasets.

In contrast, $\text{CORE}_\text{Qwen}$ demonstrates superior generalization capabilities, consistently outperforming all baselines across all four held-out datasets. 
Specifically, $\text{CORE}_\text{Qwen}$ achieves an accuracy of 60.3\% on NewsCLIPpings and 63.4\% on MDSM, surpassing the best-performing baseline by margins of \textbf{10.0\%} and \textbf{16.9\%}, respectively.
On average, our method improves over the second-best approach by approximately \textbf{11.4\%}. 
This significant improvement validates that the CORE framework, by focusing on the fundamental "conflict" logic rather than dataset-specific artifacts, successfully equips MLLMs with a more abstract and robust reasoning capability suitable for open-world multimodal misinformation detection.

\begin{table}[h]
\centering
\setlength{\tabcolsep}{4pt} 
\caption{
Cross-Dataset Zero-Shot Performance (ACC \%). Models are trained on a mixed subset (3k samples) of three datasets and evaluated zero-shot on the held-out fourth dataset.
}
\begin{tabular}{l cccc}
\toprule
\textbf{Method} & \textbf{NewsCLIPpings} & \textbf{DGM$^4$} & \textbf{MMFakeBench} & \textbf{MDSM} \\
\midrule

HAMMER & 50.3 & 47.2 & 54.1 & 46.5 \\
\midrule 
FKA-Owl & 43.8 & 46.7 & 53.3 & 42.9 \\
AMD & 40.2 & 42.1 & 46.1 & 40.3 \\
\midrule 

\rowcolor{LightGreen}
\textbf{$\text{CORE}_\text{Qwen}$} & \textbf{60.3} & \textbf{57.3} & \textbf{62.6} & \textbf{63.4} \\
\bottomrule
\end{tabular}
\small
\label{tab:cross_dataset_results}
\vspace{-0.5cm}
\end{table}

\section{Prompting vs. Training}
\label{sec:prompt_vs_training}

A natural question arises regarding the contribution of the proposed training framework: \textit{Can MLLMs achieve similar conflict detection capabilities by simply using a similar prompting strategy without the specialized training?} 

To answer this, we conducted an experiment where we directly prompted the MLLMs. Specifically, we instructed the models to first generate the conflict factors ($C_1$ and $C_2$) based on the input image and text, and subsequently use these factors to deduce whether the news is real or fake. We applied this prompting strategy to the backbone model Qwen2.5-VL-3B, as well as to two significantly larger and more powerful state-of-the-art MLLMs: Llama-3.2-Vision-90B and Seed-1.6.

We compared these "Prompt-only" baselines against our $\text{CORE}_\text{Qwen}$ fine-tuned on 100 samples (Rapid Adaptation). The results are reported in Table~\ref{tab:prompt_vs_train}.

\begin{table}[h]
\centering
\setlength{\tabcolsep}{4pt}
\caption{
Performance comparison between direct prompting and CORE training (100 samples). \textbf{Prompt-only} denotes using the model directly with conflict-oriented instructions without MBPT/CPT training. $\text{CORE}_\text{Qwen}$ trains on 100 samples.
}
\begin{tabular}{l cccc}
\toprule
\textbf{Method} & \textbf{DGM$^4$} & \textbf{MDSM} & \textbf{MMFakeBench} & \textbf{NewsCLIPpings} \\
\midrule

Qwen2.5-VL-3B & 45.3 & 46.2 & 55.3 & 47.1 \\
Llama-3.2-Vision-90B & 49.7 & 53.5 & 53.8 & 58.6 \\
Seed-VL-1.6 & 52.1 & 59.3 & 62.1 & 64.7 \\
\midrule

\rowcolor{LightGreen}
\textbf{CORE$_{\text{Qwen}}$} & \textbf{59.7} & \textbf{69.0} & \textbf{73.5} & \textbf{64.3} \\
\bottomrule
\end{tabular}
\small
\label{tab:prompt_vs_train}
\end{table}

\textbf{Analysis.} As shown in Table~\ref{tab:prompt_vs_train}, directly prompting the models yields suboptimal results compared to the CORE framework, even when using significantly larger models. 
The prompt-only Qwen2.5-VL-3B achieves only 45.3\% on DGM$^4$ and 46.2\% on MDSM. In contrast, by training on merely 100 samples, $\text{CORE}_\text{Qwen}$ boosts performance to 59.7\% (+14.4\%) and 69.0\% (+22.8\%) respectively. This indicates that without the explicit MBPT and CPT, the model struggles to accurately ground visual concepts and identify subtle contradictions, often hallucinating conflicts or failing to align the visual and textual modalities effectively.

In conclusion, simply instructing an MLLM to "find conflicts" is insufficient. The CORE framework is essential to endow the model with the actual capability to perceive and reason about these conflicts, achieving superior generalization with minimal data.

\section{Data Leakage and Overlap Analysis}
\label{sec:appendix_data_overlap}

To ensure that the reported performance gains reflect genuine generalization rather than dataset bias or memorization, we explicitly clarify the relationship between our training sources (FineHARD and SAMM) and the evaluation benchmarks (NewsCLIPpings, DGM$^4$, MMFakeBench, and MDSM). We conduct both a source-based qualitative analysis and a rigorous CLIP-based empirical verification to rule out data leakage.

\subsection{Source and Distribution Analysis}
\textbf{1. Pre-training Data (FineHARD vs. Benchmarks):} The FineHARD dataset, used for our pre-training, is constructed based on LAION-2B. This dataset predominantly consists of general-domain natural images and web-crawled captions. In contrast, the evaluation benchmarks (e.g., NewsCLIPpings and DGM$^4$) are derived primarily from the VisualNews dataset, which focuses strictly on news events and journalistic imagery. There is a fundamental domain gap between the general "in-the-wild" distribution of LAION-2B and the specific "news-caption" distribution of the benchmarks, minimizing the likelihood of direct overlap.

\textbf{2. CAC (SAMM vs. Benchmarks):} The SAMM dataset, utilized for the Conflict-Aware Contrastive (CAC) learning, employs a distinct generation pipeline for creating manipulations. The benchmarks (e.g., DGM$^4$ and NewsCLIPpings) rely on manipulation techniques that differ significantly from the patterns in SAMM (See Appendix \ref{appendix-bench}). Consequently, there is no overlap in the manipulation logic or the specific samples used.

\subsection{Empirical Verification via CLIP Similarity}
To further quantitatively verify the absence of overlap, we conducted a comprehensive similarity search across the datasets.

\textbf{Methodology.} We employed a pre-trained CLIP model to extract features from:
\begin{itemize}
    \item The specific subsets of the FineHARD (Used in MBPT) and SAMM (CAC) datasets that are actually utilized for training.
    \item The full test sets of all four benchmarks: NewsCLIPpings, DGM$^4$, MMFakeBench, and MDSM.
\end{itemize}
For every pair of samples $(S_\text{train}, S_\text{test})$—where $S_\text{train}$ is from the training sources and $S_\text{test}$ is from the benchmarks—we calculated a composite similarity score $\text{Score}_\text{sim}$. This score is defined as the sum of four cross-modal and uni-modal cosine similarities:
\begin{equation}
    \text{Score}_\text{sim} = \text{Sim}(I_\text{train}, I_\text{test}) + \text{Sim}(T_\text{train}, T_\text{test}) + \text{Sim}(I_\text{train}, T_\text{test}) + \text{Sim}(T_\text{train}, I_\text{test})
\end{equation}
where $I$ and $T$ represent the image and text embeddings, respectively.

\textbf{Results.} We identified and retrieved the top-200 pairs with the highest $\text{Score}_\text{sim}$ from the millions of potential combinations. A manual inspection of these top-200 pairs was conducted. The inspection revealed that even among the pairs with the highest similarity scores, there were no identical images or captions, and no content duplication was observed. The matches were primarily based on broad semantic similarities (e.g., two different images containing a "dog" or a "politician") rather than data leakage.

\textbf{Conclusion.} Both the source provenance analysis and the empirical feature matching confirm that there is no overlap between our training data (FineHARD, SAMM) and the evaluation benchmarks. The performance improvements reported in this paper are therefore attributed to the model's robust reasoning capabilities rather than memorization of testing samples.

\end{document}